\documentclass[10pt,twocolumn,letterpaper]{article}

\usepackage[pagenumbers]{iccv} % To force page numbers, e.g. for an arXiv version

\newcommand{\ours}{VITAL\xspace}

\usepackage{algorithm}
\usepackage{algpseudocode}
\usepackage{multirow} 
\usepackage{arydshln}
\usepackage[dvipsnames]{xcolor}
\usepackage{subcaption}
\usepackage{stix}
\usepackage{pifont}
\newcommand{\myparagraph}[1]{\vspace{2pt}\noindent{\bf #1}}

\usepackage{amsmath}

\definecolor{arrow_color}{RGB}{145,0,10}
\definecolor{activation_color}{RGB}{0,77,151}
\usepackage{placeins} 
\usepackage[accsupp]{axessibility}  % Improves PDF readability for those with disabilities.

\definecolor{iccvblue}{rgb}{0.21,0.49,0.74}
\usepackage[pagebackref,breaklinks,colorlinks,allcolors=iccvblue]{hyperref}

\def\paperID{6726} % *** Enter the Paper ID here
\def\confName{ICCV}
\def\confYear{2025}

\title{\ours: More Understandable Feature Visualization through Distribution Alignment and Relevant Information Flow}

\author{
    Ada Görgün \quad Bernt Schiele \quad Jonas Fischer \\ 
    {\tt\small \{agoerguen, schiele, jonas.fischer\}@mpi-inf.mpg.de} \\
    Max Planck Institute for Informatics, Saarland Informatics Campus, Germany
}

\begin{document}
% \maketitle
% Place figure before the abstract
\twocolumn[{%
    \maketitle
    \begin{center}
        \includegraphics[width=\linewidth]{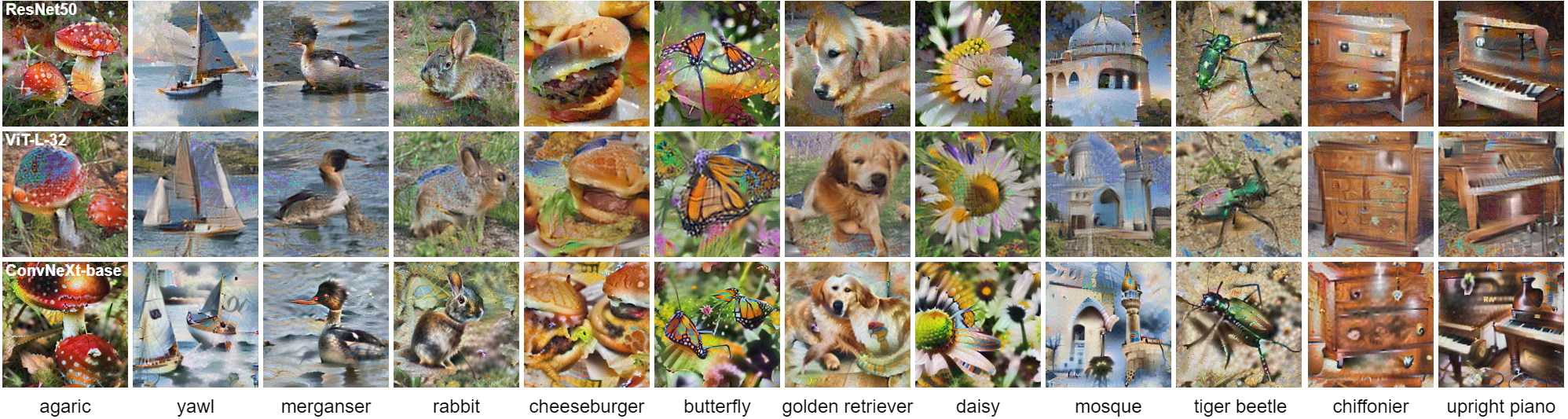}
        \captionof{figure}{\textit{Feature visualization through distribution matching.} Unlike traditional feature visualization (FV) methods, which often produce artifacts or repetitive patterns, \ours generates more understandable visualizations. Our approach scales effectively to modern architectures (rows), generalizes well across diverse classes (columns), and better captures meaningful network representations. }
        \label{fig:fig_teaser}
    \end{center}
    % \vspace{2mm}  % Add space to separate figure from text
}]

% \bernt{thanks for the teaser figure! as suggested by Jonas I would put the most striking ones in the front - such as agaric, yawl, merganser, rabbit, cheeseburger }

% \noindent
\begin{abstract}
%Ensuring the transparency and interpretability of neural networks is critical in high-stakes applications, such as medical imaging and autonomous driving. In this context, feature visualization serves a powerful tool for understanding the decision-making processes of deep neural networks when complemented with their discovered mechanistic circuits. In feature visualization, we generate synthetic images that elicit a strong response for neurons of interest and this allows one to visually inspect and consequently understand the features that are encoded in a network. 
Neural networks are widely adopted to solve complex and challenging tasks. Especially in high-stakes decision-making, understanding their reasoning process is crucial, yet proves challenging for modern deep networks. Feature visualization (FV) is a powerful tool to decode what information neurons are responding to and hence to better understand the reasoning behind such networks. In particular, in FV we generate human-understandable images that reflect the information detected by neurons of interest. However, current methods often yield unrecognizable visualizations, exhibiting repetitive patterns and visual artifacts that are hard to understand for a human.
%Besides, their scalability to deeper architectures is unfortunately limited.
To address these problems, we propose to guide FV through \textbf{statistics of real image features} combined with measures of \textbf{relevant network flow} to generate prototypical images. Our approach yields human-understandable visualizations that both qualitatively and quantitatively improve over state-of-the-art FVs across various architectures. As such, it can be used to decode \textbf{which} information the network uses, complementing mechanistic circuits that identify \textbf{where} it is encoded.\!\footnote{Code is available at: \href{https://github.com/adagorgun/VITAL}{\textnormal{VITAL Framework}}}

% As such, it can be used to decode \textbf{which} information is used by the network's reasoning process, complementing the methodology of mechanistic circuits that identify \textbf{where} relevant information is encoded.

% Code is available at: \href{https://github.com/adagorgun/VITAL}{\textnormal{VITAL Framework}}
% \bernt{do we really want to have the last sentence in the abstract? it is already in the first paragraph of the intro and that seems the better place for me?} 

\end{abstract}

\section{Introduction}
\label{sec:intro}

Deep neural networks have achieved remarkable success across science and industry. Their reasoning process is, however, inherently opaque. Different lines of research in Explainability of Machine Learning have been addressing these issues as understanding of how a neural network arrives at a decision is crucial, especially in safety-critical domains. To understand which input features are used for a decision, both post-hoc, as well as built-in methodologies have been proposed, which give information about input-output relations in the form of attribution or relevance scores~\cite{LRP,SmoothGrad,InteGrad,Bcos}.
Yet, these offer only a partial understanding of a model's decision-making process.
In the scope of \textit{Mechanistic Interpretability}~\cite{mechanisticdisc, mechanisticrev}, experts are interested in finding explanations that uncover what the model \textit{internally} learns.
In this line of research, methods that discover paths of information flow -- so-called circuits -- have been proposed \cite{pmlr-v139-fischer21b,NEURIPS2023_34e1dbe9, moakhar2024spadesparsityguideddebuggingdeep}, along with approaches to discover network components, such as groups of neurons, that together encode a concept \cite{CRAFT}. While these methods reveal \textit{where} information is propagated towards a particular decision, they fall short in identifying \textit{which} information is encoded.
Feature visualization (FV) bridges this gap by generating images that correspond to what triggers a strong response in a given neuron or group of neurons. This is traditionally achieved by optimizing for an image that maximizes the response of the selected neuron(s) through gradient ascent. Complemented with circuits, FV thus help represent the internal reasoning process in human-understandable terms.

Feature Visualizations have been an integral part of generating explanations for deep neural networks, adopted early on for CNN models~\cite{Simonyan2013DeepIC}. FVs for earlier architectures~\cite{Inception,VGG} through explicit regularization \cite{Vedaldi2015,Nguyen2015,Bilateral} or transformations \cite{Mordvintsev2015InceptionismGD,olah2017feature} promote robustness against variations within the image.
Already in these early works, irrelevant features as well as artificial, repetitive patterns rendered interpreting the visualizations challenging.
For subsequent, larger networks, new methods had to be developed and further refined to get towards realistic image visualizations~\cite{multifaceted,yin2020dreaming,Fel2023}, incorporating elaborate regularizations or reparametrizations in the Fourier space to dampen high-frequency features. Still, newer methods also struggle with repetitive patterns and irrelevant features on modern architectures (see Fig.~\ref{fig:fig_class_neurons}) and show deteriorating interpretability when scaling to larger models.

% \noindent
In this work, we suggest to re-frame FV in terms of optimizing an \textit{alignment to realistic feature distributions} instead of \textit{activation maximization}.
In particular, we consider the per-channel distribution of activations in intermediate layers across a set of reference images (e.g., training images of a class, top-activating images for a neuron) and optimize our feature visualization to follow a similar distribution of activations. As such, repetitive patterns -- visible as a large number of high activations in a channel -- are discouraged unless natural in the original data.
Similarly, artificial features or colors, which might highly activate a neuron but are far from the input manifold, are discouraged (see \cref{fig:fig_teaser}).
By further incorporating \textit{feature relevance} for FVs of neurons for adjusting the feature distributions by how relevant they are to the neuron under study, we further remove irrelevant features from the visualizations that impacted the faithfulness of existing FVs.
We provide an efficient implementation of our approach, \ours, making use of a distribution matching algorithm through which we can back-propagate. This enables \ours to seamlessly scale to modern architectures, including large networks and Vision Transformers (ViTs),
% \bernt{english is off here - what do you want to say? maybe `it runs as fast as existing approaches'. Not sure if we want to state this in the intro? The key is to argue imho that the approach scales to current deep neural network architectuers. unclear to me if we want say anything about `speed' here?}
but yields more interpretable FVs, which we show in a set of experiments including a human user study.
Our \textbf{contributions} are:
\begin{itemize}
    \item we propose a novel method, \ours, to optimize feature visualizations through alignment with feature distributions of real data instead of maximizing activations,
    \item we propose to incorporate feature relevance scores in the optimization to focus on exactly those features that are perceived by the target neuron from the feature visualization, and
%    \item we suggest an efficient algorithm to match feature distribution (TODO backprop,)
    \item we show both qualitatively and quantitatively on a diverse set of metrics -- including a human user study -- that \ours yields more understandable and accurate descriptions of what information is encoded in a neuron.
\end{itemize}

% \bernt{maybe replace `realistic' with `understandable'? - would fit better to the story/title and it is not so clear if our iamges are actually realistic...} 
\section{Related Work}
\label{sec:related_work}
\textbf{Mechanistic Interpretability.}
Understanding the internal decision-making process of neural networks has regained significant attention as part of \textit{Mechanistic Interpretability (MI)} of neural networks.
While definitions of MI can differ~\cite{mechanisticdisc}, the essence is to find explanations of neural networks that allow an interpretation of how \textit{internal components}, such as neurons or layers, operate together to give reason to a prediction. We refer to recent reviews for a comprehensive summary~\cite{mechanisticrev, mechanisticdisc}.
One key area of research is the discovery of circuits, which describe along which network paths information is propagated to a target neuron.
We here discuss how to describe \textit{which} information is encoded in neurons through feature visualization, which also has shown promising results in explaining the information encoded in circuits or feature directions~\cite{pmlr-v139-fischer21b, CRAFT}.

\myparagraph{Model Inversion.} Model inversion techniques reconstruct input images from model outputs by optimizing images to match predictions, revealing learned representations and privacy risks. Recent works have improved inversion fidelity through data augmentations \cite{pmlr-v162-ghiasi22a}, distribution matching \cite{yin2020dreaming}, reinforcement learning \cite{ReinforcementAttack}, and generative networks \cite{yu2023generator}. Unlike model inversion, which visualizes class-specific information only, FVs can capture information of \textit{any} network component, including individual neurons, neuron combinations, and channels, to better understand network representations.

\myparagraph{Feature Visualization.} Feature visualization techniques aim to generate an image that maximizes the activation of specific network parts (e.g., a neuron)~\cite{Simonyan2013DeepIC, Zeiler2014}. To yield understandable visualizations, researchers suggested to impose constraints based on local variance \cite{Vedaldi2015}, apply blurring \cite{Nguyen2015, Bilateral}, or ensure consistency across transformations~\cite{Mordvintsev2015InceptionismGD,multifaceted} to promote robustness. One of the most widely used approaches is called preconditioning \cite{olah2017feature}, where optimization is performed on the Fourier basis, ensuring that data is decorrelated and whitened. Most recently, MACO \cite{Fel2023} improved this preconditioning, suggesting to only optimize the phase while fixing the magnitude to that of a typical image.
%Additionally, inspired by SmoothGrad \cite{SmoothGrad}, they reformulate the transparency mapping introduced in \cite{mordvintsev2018differentiable} to identify the most affected regions during optimization. 
Unfortunately, good visualizations for modern architectures such as large ResNets~\cite{He2015DeepRL} are still hard to achieve. To address this challenge, statistically learned priors based on generative models such as GANs \cite{FV_GAN, yu2023generator}, or autoencoders \cite{FV_Autoencoder,Wang2022TraditionalCN} are leveraged. A key drawback of these, compared to traditional FV, is that it is unclear what of the visualization can be attributed to the neural network under study and what is due to the generative model. 

\section{\ours Framework}
\label{sec:method}
%In this section, we introduce our VITAL framework with necessary preliminaries in section \ref{subsec:pre}. Then, in section \ref{subsec:lrp}, we give a brief overview of Layer-wise Relevance Propagation (LRP) that incorporates the information about the network flow, followed by our Sort-Matching (SM) loss in section \ref{subsec:sm}. In the subsequent sections \ref{subsec:class_neuron} and \ref{subsec:inner_neuron}, we describe how to effectively visualize intermediate neurons and class neurons, respectively. Finally, in \ref{subsec:diversity}, we briefly explain how we deal with polysemantic neurons.
Existing FV methods suffer from overly repetitive patterns, artificial, and irrelevant features.
% To tackle these issues, we suggest a new paradigm focused on optimizing an alignment between features of the learned image and features of real images.
% By using distributions of activations across different layers, we ensure that both coarse- as well as fine-grained information in our FV corresponds to a typical input (sec.~\ref{sec:vital:matching}).
To address this, we propose aligning feature distributions between learned and real images across layers, capturing both coarse and fine-grained information (\cref{sec:vital:matching}). 
This discourages the use of artificial
features and repetitive patterns, which typically spike the activation in classic approaches.
We further suggest to take into account relevance scores, matching the distributions of activations weighted by their relevance to the target, which discourages features in the FV that are irrelevant for the target neuron (sec.~\ref{sec:vital:relevance}).
This becomes particularly important for the visualization of inner neurons. For example, while grass should be irrelevant for a bird's beak, it might still co-activate channels across multiple bird images, which misleads interpretations. Incorporating LRP eliminates irrelevant correlations in activation maps, ensuring that only the features contributing to the target neuron’s activation are visualized. We provide an example in Fig.~\ref{fig:fig_method_overview} and further details about our approach in Appendix \cref{sec:supp_method}.

\begin{figure}
    \centering
    \centerline{\includegraphics[width=.97\linewidth]
    {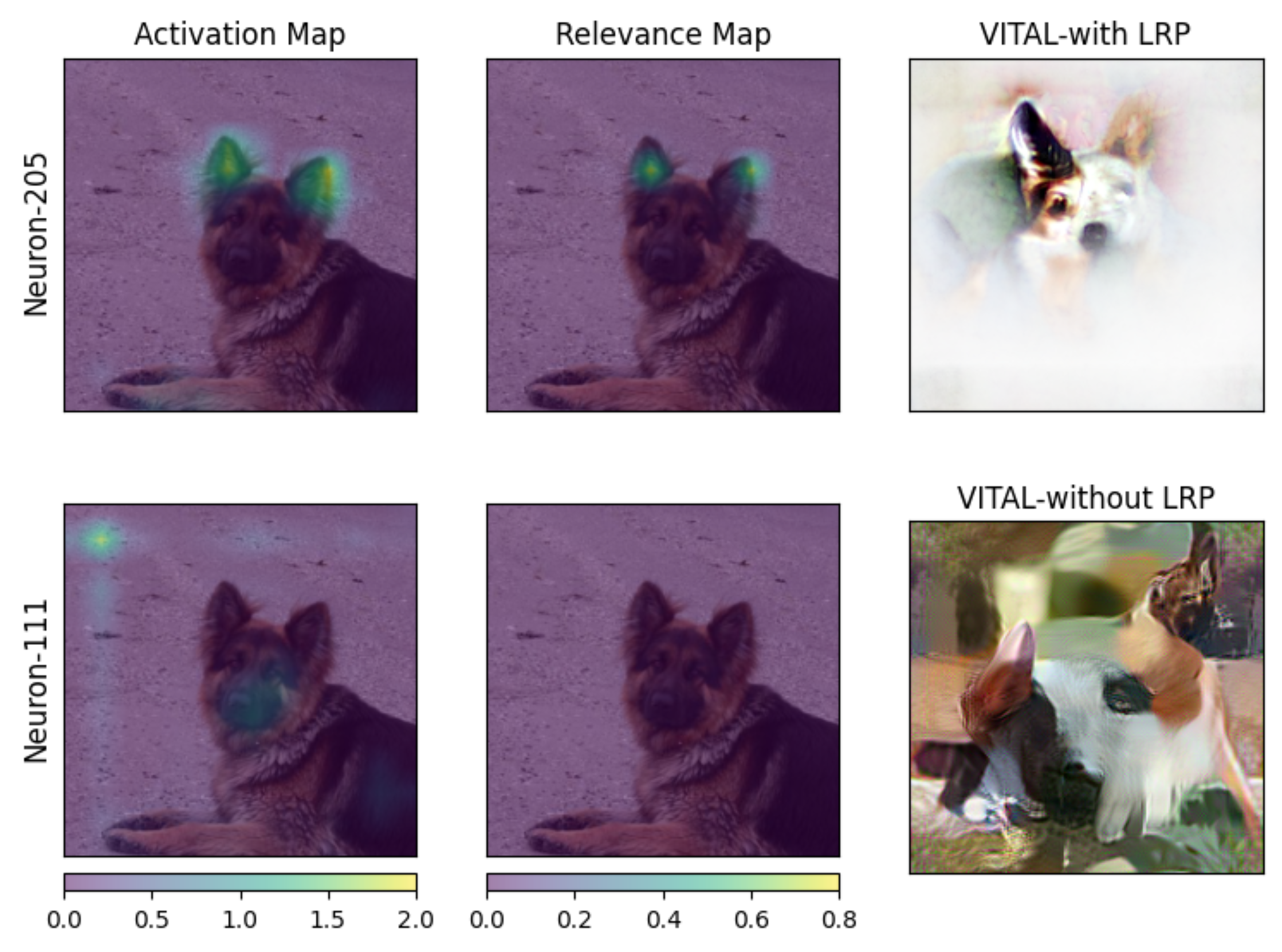}}
    \caption{\textit{Importance of relevance scores.} Given an image selected by its relevance to a neuron detecting an ear, we apply LRP from this neuron back to a preceding building block. We show the activation  \textbf{(first column)} and relevance maps \textbf{(second column)} of two neurons from this block, respectively. While the activation is high for the ears (top) and background (bottom),  only the ear is relevant for the target neuron. The difference between \ours with and without LRP  \textbf{(third column)} shows that incorporating relevance helps to avoid visualizing irrelevant features.}
    \label{fig:fig_method_overview}

\end{figure}

%\bernt{the above paragraph does not really give an overview/idea of the approach - it just lists the (known) components described in the following - it lacks \textit{excitement} and \textit{novelty} -- I am coming back to the argument that we need a clear list of contributions to know what we are targeting in the paper (intro, related work, methods, experiments).}
%-------------------------------------------------------------------------
\subsection{Notation}
\label{subsec:pre}
We consider a neural network $f(\cdot; \theta) : \mathcal{X} \rightarrow \mathcal{Y}$ as a composition of $L$ functions corresponding to layers, i.e., $f = f^{(L)} \circ f^{(L-1)} \circ \dots \circ f^{(2)} \circ f^{(1)}$ from an input space $ \mathcal{X} \in \mathbb{R}^{C_0 \times H \times W} $ to an output space $ \mathcal{Y} \in \mathbb{R}^{c}$. Here, $\theta$ are the (trained) network parameters, $C_0$ is the number of channels (typically $1$ or $3$), and $c$ is the number of classes. 
For ease of notation, we will use $f^{(l)}(x) = f^{(l)} \circ \dots \circ f^{(2)} \circ f^{(1)} (x),$ to indicate the network output of layer $l$ on input $x\in\mathcal{X}$.
Without loss of generality, we denote the feature space of layer $l$ flattened over its spatial dimensions as $A^{(l)}(x)=f^{(l)}(x) \in \mathbb{R}^{C_l \times D} $, where $C_l$ is the number of channels.
%For simplicity, we denote the activation space of the generated and real images as $A_{gen}$ and $A_{real}$, respectively, and omit the subscript $l$ when clear from context.

%-------------------------------------------------------------------------
\subsection{Feature Distribution Matching}\label{sec:vital:matching}

\begin{figure}
    \centering
    \centerline{\includegraphics[width=.85\linewidth]
    {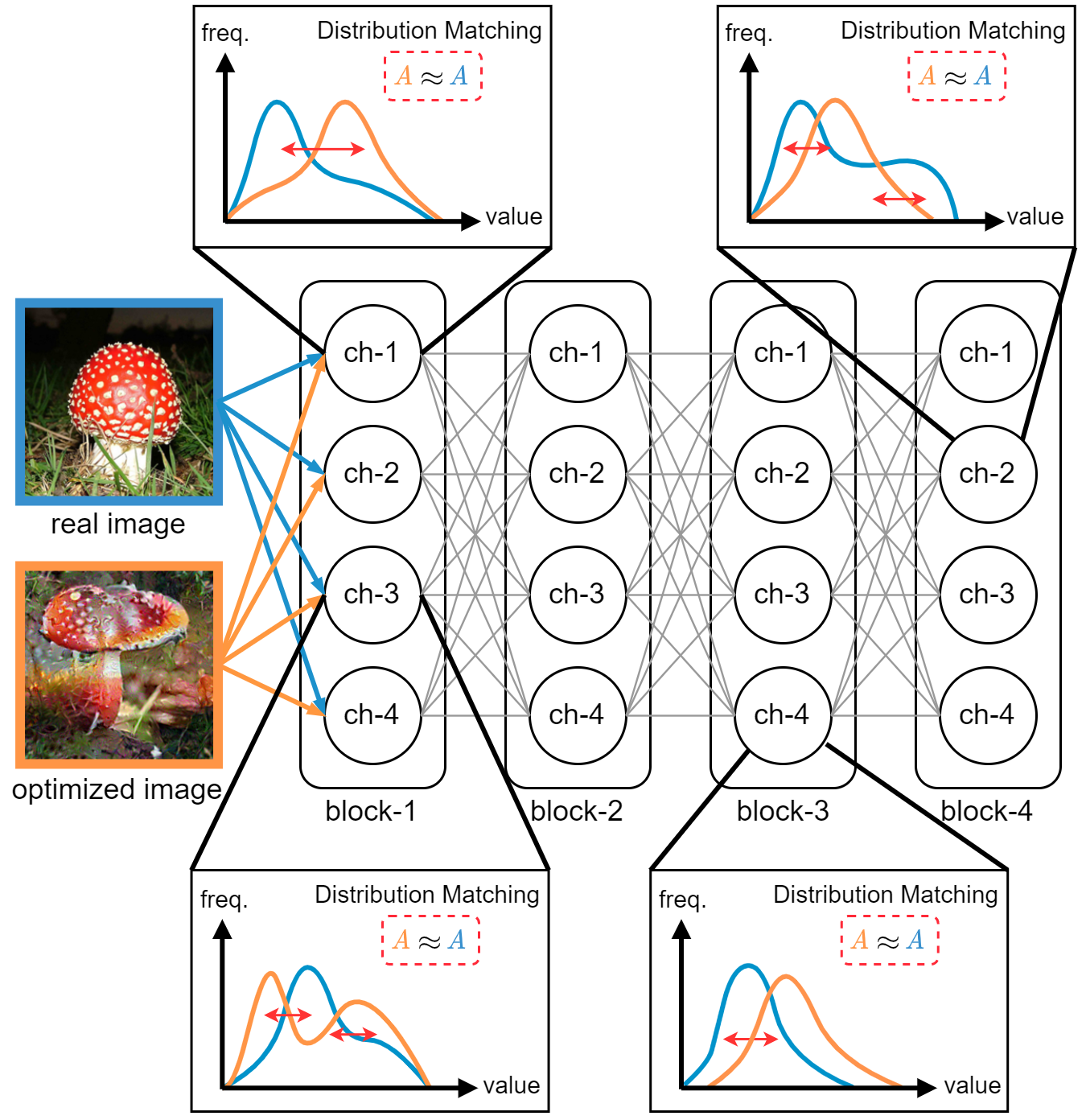}}
    \caption{\textit{Overview for feature distribution matching.} Activations or activation $\times$ relevance scores ($A$) are extracted from channels across network blocks for real and optimized images. For feature visualization,
    the distribution of the generated image \textcolor{orange}{$x^*$} is optimized to match those of the real reference images \textcolor{blue}{$x' \in \mathcal{X}_{ref}$}.}
    \label{fig:fig_method_diagram}

\end{figure}

Given a target of interest in the network $f$, e.g. the $i$-th neuron in layer $l$, $f^{(l)}_i$, we are after an input $x^*$ that captures the information that $f^{(l)}_i$ responds to.
Instead of finding an image that maximizes this activation,
%$x^*=\argmax_{x} f^{(l)}_i (x)$, 
as typical in the literature, we consider the distribution of features $A^{(l')}$ at layers $l'<l$ of both our generated input $x$ and inputs $x' \in \mathcal{X}_{ref}$ from reference data.
Such references could be training data for which $f^{(l)}_i$ responds highly. We would like to have similar feature distributions of $x$ compared to these reference images $x'$.
More formally, we would like to find an image $x^*$ such that $\forall l'<l,x'\in\mathcal{X}_{ref}:$
\begin{equation}\label{eq:distapprox}
\text{dist}\left(A^{(l')}(x^*)\right) \approx \text{dist}\left(A^{(l')}(x')\right)\!,
\end{equation}
where $\text{dist}(\cdot)$ indicates the distribution of the activations. In \cref{fig:fig_method_diagram}, we illustrate the alignment of feature distributions for sample channels and blocks.

Investigating empirical distributions of activations, we observe that they do not consistently follow a typical prior, such as a Gaussian, and hence decided to match empirical distributions directly.
In particular, we leverage a result on sort-matching from the style-transfer literature~\cite{SM}, which allows us to compute a matching between two feature vectors, here $A^{(l')}(x^*)$ and $A^{(l')}(x')$, that we can \textit{back-propagate} through. We can, hence, efficiently compute gradients with respect to $x^*$.
Accumulating the difference between the (sorted) feature vectors of our current generated image $x^*$ and the (sorted) feature vectors for all reference images, and across layers defines a \textit{loss function} that reflects Statement~\ref{eq:distapprox}.
So, optimizing this loss with respect to the generated image $x^*$, improves the alignment of the features of our generated image with real images.
%We next recapitulate the key idea behind the differentiable sort-matching procedure before discussing the selection of reference images and how to incorporate relevance scores to focus on non-spurious features.

\subsubsection{Sort-Matching Loss}
\label{subsec:sm}
The key idea behind the sort-matching loss suggested by \citet{SM} is to compute the difference of the distribution of two arrays, $z=A^{(l')}(x^*)$ and $z'=A^{(l')}(x')$, by sorting these two arrays that allow for differentiation to $x^*$.
In particular, we first compute indices $\pi$ and $\pi'$ that sort the arrays $z$ and $z'$, respectively. We then compute an indexing $\Bar{\pi}, \Bar{\pi}'$ that sorts these \textit{indices}. Intriguingly, applying both indices yields the original order, i.e., $z = \Bar{\pi}(\pi(z))$.
We can now match $z$, which is a function of  $x^*$, and  $z^r=\Bar{\pi}(\pi'(z'))$, which reorders the sorted values of $z'$ \textit{based on the indexing from $z$}.
Computing the MSE between these two arrays
\begin{equation}\label{eq:mse}
\text{MSE}(z, z^r) = \frac{1}{|z|} \sum_{i=1}^{|z|} \left(z_i - z_i^r\right)^2\!\!\!\!,
\end{equation}
we can differentiate for $x^*$, allowing for gradient-based optimization. For multiple real feature vectors, we average the transformed $\pi'(z')$ vectors to obtain a representative prototype before applying $\Bar{\pi}$ to obtain $z^r$.
% and hence use gradient-based optimizers to improve the feature visualization.
%We provide an example computation of sort-matching in the Appendix.

%Given two feature vectors $z \subseteq A_{gen}$, and $\hat{z} \subseteq A_{real}$ of dimension $D$, we formalize our histogram matching problem as transforming $z$ into $z^*$ that matches the values of reference $\hat{z}$. With Sort-Matching (SM), we aim to generate an output $z^*$ by matching the sorted values of $z$ and $\hat{z}$. Yet, this cannot be applied to our scenario when $z$ is learnable as sorting is a discrete assignment problem. Inspired by recent work from style-transfer \cite{SM}, we propose a transformation on $\hat{z}$ as illustrated in Fig. \ref{fig:fig_method_overview}-b, allowing $z$ to remain unchanged during the distribution matching operation. The matching is done by minimizing $\mathcal{L}_{SM}$, which is the MSE loss between $z$ and transformed $\hat{z}$ during optimization. This loss can be easily extended to multiple channels and multiple building blocks of our network. For the case where we have multiple real feature vectors, we simply average the transformed $\hat{z}$ vectors and obtain their representative prototype to be used in the SM loss. In our method, we use all the channels of the building blocks during matching.

\subsubsection{Selection of Reference Images}
\label{subsec:selection}

%We define our image set as the entire ImageNet training data. Considering the computational needs, the subset of the training data can also be used. After defining our image set, we divide our selection into two sub-problems for: 1) class visualizations, 2) intermediate neuron visualizations. In the first one, we select random $N$ images for feature distribution matching. In the second one, we first select sub-regions to identify specific concepts that neurons respond to within a localized context. To formalize how we select the $N$ important images,  similar to CRAFT \cite{CRAFT}, we first crop and resize the training images into patches to obtain our auxiliary dataset (see App. for more detail). Then, we obtain the activations for each patch in a batch-wise manner. At each iteration, we store and update the top-$N$ patches and eliminate the patches that are coming from the same original image. In the case where the model is a CNN, the score for each patch is formulated as the global average pooling of activations across their spatial dimension. Ultimately, we obtain the top-$N$ images to be used in the next stage of our framework based on the top-$N$ patches.
% Typically, one would select a set of images from data that is close to the distribution that the model of interest was trained for, e.g. training that the target neuron of interest is highly activated for. 

Typically, the reference set should reflect the distribution relevant to the target neuron. For class neurons, we directly use a random selection of images from that class. For intermediate neurons, we select the top-$k$ patches (from distinct images) relevant for this neuron, following \citet{CRAFT}. In brief, we crop and resize training images into patches and score each based on the global average pooled activation across spatial dimensions, keeping the top-$k$ patches from unique images. These define our reference $\mathcal{X}_{ref}$ for feature alignment. We provide additional sampling experiments, including size sensitivity, cluster-based sampling, dataset generalization, and corruption analysis, in the Appendix \cref{subsec:supp_ablation_ref_images}.

\subsection{Incorporating Relevance Scores}\label{sec:vital:relevance}
% \bernt{detail: capitalization is not consistent for sections/subsections - either we capitalize all or none\dots}
Especially for visualizations of intermediate neurons, irrelevantly activated features pose a problem. For example, a neuron detecting a facial feature of animals might be correlated with a background feature, such as grass, as this is predominant in  training -- an animal in nature. Classic approaches of activation maximization do not discourage such activations, which is why irrelevant features can appear in their visualizations.
 % \bernt{do you mean `image' or rather `activation' maximization?}
Similarly, matching of distributions of activations would encourage background features in the FV.
To overcome this, we incorporate \textit{feature relevance scores} that describe how much a feature (encoded by an intermediate neuron or channel) is relevant for the downstream target neuron. Methods such as layer-wise relevance propagation (LRP) provide meaningful relevance scores~\cite{LRP,LRP_ResNet}.
We provide an example of irrelevant background activation and its actual relevance computed by LRP for the target neuron for ResNet50 trained on ImageNet, in Fig.~\ref{fig:fig_method_overview}. 

We thus suggest matching feature distributions based on activation times relevance, i.e., match features $A^{(l')}(x) \odot R_n^{(l')}(x),$ where $R_n$ are the relevance scores for the target neuron $n$ under study.
These relevance scores give weight to each neuron (or pixel of a channel) describing how relevant the neuron is for target neuron $n$ on the given sample. They are hence of the same dimension as the activations, $\odot$ thus indicating the Hadamard product.
In principle, any attribution approach can be used as plug-in replacement to obtain relevance scores $R$. For the remainder of the paper, we use LRP, one of the most prominent attribution methods, and provide a comparative result for Guided Backpropagation \cite{guided_backprop} in Appendix \cref{subsec:supp_ablation_attribution}. Besides intermediate neurons, we also applied relevance scores to class neurons in Appendix \cref{subsec:supp_analysis_lrp_class}.

\subsection{Further Considerations}

\myparagraph{Transparency Map.}
Irrelevant areas of the generated image stay mostly unchanged during the optimization, essentially representing noise.
Analogous to \citet{Fel2023}, we suggest using transparency maps based on the importance of the image location during optimization to show relevant image parts only.
In brief, we accumulate the gradients of our loss (Eq.~\ref{eq:mse}) across each step in the optimization.
%As done in SmoothGrad \cite{SmoothGrad}, we average those gradients through the whole optimization process. 
We thus ensure the identification of the areas that have been most attended to by the network during the generation of the image.
In the Appendix \cref{subsec:supp_ablation_transparency}, we illustrate the effect of the transparency map for example visualizations.

\myparagraph{Auxiliary Regularization.}
While already working reasonably in practice, existing regularizations further reduce noise and small artifacts in the generated image.
Similar to \citet{yin2020dreaming}, we penalize the total variance (TV) and $\ell_2$ norm of our FV with parameter $\alpha_{\text{TV}}$ and $\alpha_{\ell_2}$ as in:
\begin{equation}
\mathcal{L}_{\text{aux}}(x^*) = \alpha_{\text{TV}} \mathcal{L}_{\text{TV}}(x^*) + \alpha_{\ell_2} \ell_2(x^*),
\end{equation}
where $x^*$ is the generated image.
Accordingly, the total loss to be minimized becomes
\begin{equation}
\mathcal{L}_{\text{VITAL}}\left(x^*, \mathcal{X}_{ref}\right) = \mathcal{L}_{\text{SM}}\left(x^*, \mathcal{X}_{ref}\right) + \mathcal{L}_{\text{aux}}(x^*),
\end{equation}
where $\mathcal{X}_{ref}$ are the set of real reference images, $\mathcal{L}_{\text{SM}}\left(x^*, \mathcal{X}_{ref}\right)$ is the loss term as specified in Eq.~\ref{eq:mse} summed over different layers.
We ablate on different hyperparameter settings to show their impact on the final visualization in the Appendix \cref{subsec:supp_ablation_hyp}.

\myparagraph{Choice of Layers for Alignment.}
While agnostic to the architecture, it is more efficient to focus on aligning features from only few, important layers.
To decide on which layers to consider, we conducted a small ablation study (see Appendix \cref{subsec:supp_ablation_layers}) to examine the effects of layers on the visualizations.
In a ResNet50, we find that only optimizing for feature alignment of first and last block output is sufficient to produce high-quality images, already capturing the important low- respectively high-level features in the image.

\myparagraph{Visualization of Concepts.}
As part of Mechanistic Interpretability, people are interested in finding concept-based explanations of model behavior.
These concepts might be feature directions encoded through multiple neurons in a layer, for example, discovered by CRAFT~\cite{CRAFT}. 
In \ours, we modify the initialization of relevances of target neurons in LRP to reflect the weights given by the feature direction. 
Through this modification, \ours can give \textit{meaning}, to these feature directions.
We provide technical details and results of this approach in the Appendix \cref{subsec:supp_concept_vis}.

%\myparagraph{Disentangling polysemantic features.}
%An interesting area of recent study is the problem of polysemantic neurons, which are neurons that encode different semantic meaning depending on their activation context.
%Such polysemanticity can have tremendous effect on Explainability methods, including feature visualizations.
%Knowing a polysemantic neuron, by using a simple disentangling procedure, we can then for example visualize the distinct semantic features of the 'cauliflower-alarmclock'-neuron (see Fig.~\ref{fig:fig_diversity}).
%More specifically, for a given polysemantic neuron, we find the top-5 classes that are most relevant for it using LRP. As before, we then select top-$N$ patches extracted from the training set of those classes that elicit a strong response for the selected neuron.
%Applying $k$-means clustering~\cite{KMeans} with $k$ equal to the number of concepts in the neuron yields cluster centers for the these concepts. For each center, we obtain the top-$N'$ patches that are closest to the center based on the Euclidean distance. Finally, we apply \ours to each of the sets of patches to generate disentangled visualizations. In Fig. \ref{fig:fig_diversity}, we illustrate a polysemantic neuron from MACO and their disentangled versions from \ours and provide more examples in the Appendix.

\myparagraph{Runtime.}
Assuming that feature distributions of the reference images are precomputed, 
\ours runs in time $O(|\mathcal{X}_{ref}| * L * C * P log P)$, for $|\mathcal{X}_{ref}|$ reference images, $L$ layers considered for matching, $C$ (maximum) number of channels in those layers and $P$ the (maximum) number of pixels in a feature map. As discussed before, $L$ is usually small, in the following, we use $L=5$ layers for matching.
When integrating LRP scores in our matching, we additionally have the cost of one backward pass through the model.
In practice, it takes roughly 40 seconds to generate an image \textit{including} computation of distributions for reference images. For comparison, MACO takes 23-28 seconds and DeepInversion 1-3 minutes, depending on their setting.
Similarly expensive as DeepInversion is the generation of inner neurons with \ours ($L=4$), with 2-3 minutes \textit{including} computation of distributions.% for reference images.%, which is still reasonable given that this is usually not a live task.

\begin{figure}
    \centering
    \begin{subfigure}{0.48\linewidth}
        \centering
        \includegraphics[width=.78\linewidth]{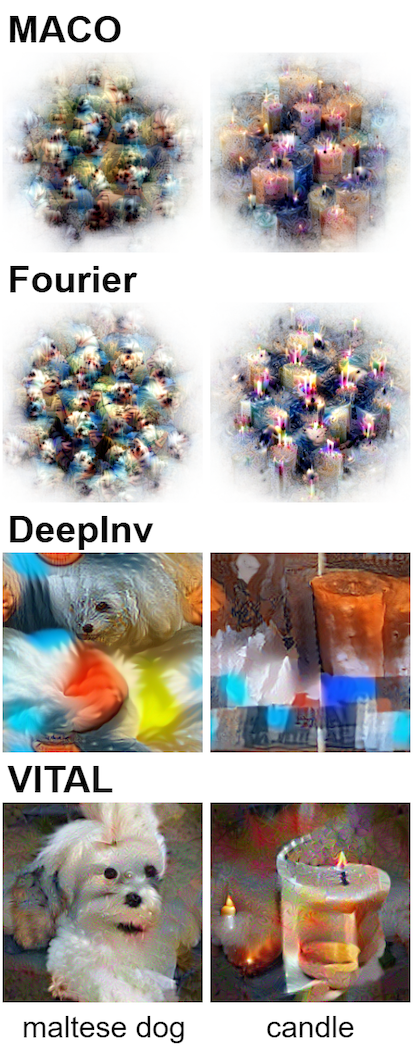}
        \caption{ResNet-50}
    \end{subfigure}
    \hfill
    \begin{subfigure}{0.48\linewidth}
        \centering
        \includegraphics[width=.78\linewidth]{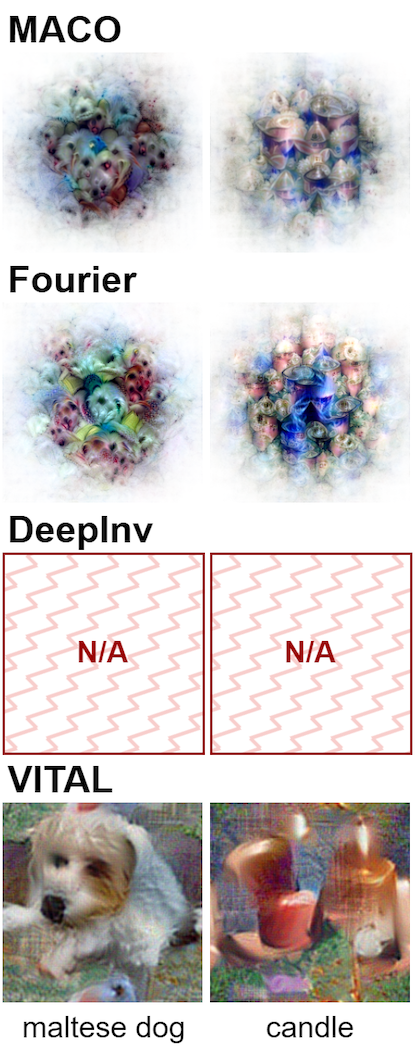}
        \caption{ViT-L-32}
    \end{subfigure}
   
    \caption{\textit{Example class visualizations.}  We show ImageNet class visualizations for (a) ResNet50 and (b) ViT-L-32.MACO and Fourier-based FV (top rows) produce repetitive, hard-to-interpret patterns. DeepInversion (3rd row) improves interpretability but introduces artifacts and lacks ViT compatibility. \ours arguably yields much more interpretable visualizations. For additional results, including failure cases, we refer to Appendix \cref{sec:supp_experiments}.}
    \label{fig:fig_class_neurons}
\end{figure}

% , yet, as all methods, has problems with complex spatial arrangements (see the ambulance)

\section{Experimental Results}
\label{sec:experiments}
% \bernt{general comment - since this paper is about feature visualization - I would argue we need far more qualitative  results in terms of feature visualization with convincing captions that argue that can gain interesting insights from those visualization. At the moment - while having two figures - I feel this is still not convincing enough and just seems to show that we are somewhat better than a few previous methods - is that convincing enogh for everyone?}

To evaluate \ours, we compare against Fourier \cite{olah2017feature} as standard baseline, DeepInversion (DeepInv) \cite{yin2020dreaming} as state-of-the-art class visualization, and MACO \cite{Fel2023}, which is generally considered state-of-the-art for both class and inner neuron visualizations.
We implemented \ours in PyTorch and compared on ResNet50 \cite{He2015DeepRL}, DenseNet121 \cite{densenet}, ConvNext-base \cite{convnext}, ViT-L-16 \cite{vit}, ViT-L-32 \cite{vit}, both qualitatively and quantitatively. All architectures were pre-trained on Imagenet. We further conducted human user studies assessing human interpretability of the resulting visualizations. We provide details on implementation and setup of competing methods with more results in the Appendix \cref{sec:supp_method,sec:supp_experiments}, respectively.

\begin{table}[t]
    \centering
    \scriptsize
    \begin{tabular}{l cccccc}
        \toprule
          & \multirow{2}{*}{Method} & \multirow{2}{*}{Setup} &Acc. &  \multirow{2}{*}{FID ($\downarrow$)}& \multicolumn{2}{c}{Zero-Shot Prediction}   \\ 
         \cmidrule{6-7}%\cmidrule{6}
         & & & Top1 ($\uparrow$) & & Top1 ($\uparrow$) & Top5 ($\uparrow$) \\
         \midrule
         \multirow{7}{*}{\rotatebox{90}{ResNet50}} &\textit{ImageNet} &  & - & -&  69.11 & 92.23 \\
         \cmidrule{2-7}
         & MACO & r: 224 & 29.43 & 360.74 & 12.87 & 29.73 \\
         \cdashline{2-7}\noalign{\vskip 1.0ex}
         &Fourier & r: 224 & 21.30 & 422.44 & 6.73 & 18.27  \\
         \cdashline{2-7}\noalign{\vskip 1.0ex}
         &DeepInv& bs: 64 & \textbf{100.00} & \textbf{35.76} & \underline{29.90} & \underline{55.20} \\
         \cdashline{2-7}\noalign{\vskip 1.0ex}
         &\textbf{VITAL} &  & \underline{99.90} & \underline{58.79} & \textbf{66.62} & \textbf{92.56}  \\
         \midrule

         % CONVNEXT
         \multirow{6}{*}{\rotatebox{90}{ConvNeXt}}&\textit{ImageNet} &  & - & -& 65.66 & 89.80 \\
         \cmidrule{2-7}
         &MACO & r: 224 & \underline{66.07}  & 62.55 & \underline{7.20} & \underline{19.77} \\
         \cdashline{2-7}\noalign{\vskip 1.0ex}
         &Fourier & r: 224 & 60.07 & \underline{59.60} & 2.77 & 8.30 \\
         \cdashline{2-7}\noalign{\vskip 1.0ex}
         &\textbf{VITAL} & &\textbf{99.97} &\textbf{3.92} & \textbf{63.53} & \textbf{90.30}  \\
         \midrule

         % DENSENET
         \multirow{7}{*}{\rotatebox{90}{DenseNet121}}&\textit{ImageNet} &  & - & -& 70.64 & 93.16 \\
         \cmidrule{2-7}
         &MACO & r: 224 & 9.20 & 1.80 & 9.33 & 23.20 \\
         \cdashline{2-7}\noalign{\vskip 1.0ex}
         &Fourier & r: 224 & 15.53 & 1.63 & 4.87 & 12.17  \\
         \cdashline{2-7}\noalign{\vskip 1.0ex}
         &DeepInv& bs: 64 & \textbf{100.00} & \textbf{0.20} & \underline{10.00} & \underline{25.47} \\
         \cdashline{2-7}\noalign{\vskip 1.0ex}
         &\textbf{VITAL} & & \underline{99.93} & \underline{0.27} & \textbf{58.70} & \textbf{86.93 } \\
         \midrule

         % ViT-L-16
         \multirow{6}{*}{\rotatebox{90}{ViT-L-16}}&\textit{ImageNet} &  & - & -& 64.78  &  89.31\\
         \cmidrule{2-7}
         &MACO & r: 224 & \underline{44.33} & \underline{946.96} & \underline{3.93} & \underline{10.57} \\
         \cdashline{2-7}\noalign{\vskip 1.0ex}
         &Fourier & r: 224 & 25.30 & 990.51 & 1.67 & 5.13  \\
         \cdashline{2-7}\noalign{\vskip 1.0ex}
         &\textbf{VITAL} & & \textbf{99.80} & \textbf{126.29} & \textbf{68.17 }& \textbf{92.80}   \\
         \midrule

         % ViT-L-32
         \multirow{6}{*}{\rotatebox{90}{ViT-L-32}}&\textit{ImageNet} &  & - & -& 65.83  & 90.03 \\
         \cmidrule{2-7}
         &MACO & r: 224 & \underline{24.87} & 2318.90 & \underline{17.53} & \underline{37.23} \\
         \cdashline{2-7}\noalign{\vskip 1.0ex}
         &Fourier & r: 224 & 17.03 & \underline{1983.09} & 10.30 & 28.10  \\
         \cdashline{2-7}\noalign{\vskip 1.0ex}
         &\textbf{VITAL} & & \textbf{89.60} & \textbf{147.33} & \textbf{55.97} & \textbf{85.47}   \\

         \bottomrule
    \end{tabular}
    \caption{Comparison of methods on different architectures trained on Imagenet. We provide FID scores, CLIP Zero-shot prediction scores, and top-1 classification accuracy, indicating the \textbf{best} and \underline{second best}. In the settings, "r" indicates the resolution of the visualization and "bs" is the used batch size.}
    \label{tab:combined_new}
\end{table}

\subsection{Qualitative Results}

Visually, there is a stark difference between class visualizations of \ours and existing work (see Fig.~\ref{fig:fig_class_neurons} for ResNet50 and ViT-L-32 and Appendix \cref{sec:supp_experiments} for all architectures). Where most existing methods, including MACO and Fourier-based activation maximization, show highly repetitive patterns, our approach provides much cleaner, more understandable representations of the class. While DeepInversion shows less repetition of patterns, there are often irrelevant background features, as in the case of the candle, or visual artifacts, as the blue and yellow color for the dog, dominating the image.
Yet, we acknowledge that none of the methods is flawless: All approaches, including \ours, show difficulties with complex spatial arrangements (Appendix \cref{subsec:supp_failure}).
Still, it is usually understandable what class is represented by \ours. This becomes even more evident when comparing methods across more complex architectures (e.g., ViT, ConvNeXt), where most approaches struggle to remain interpretable. In contrast, \ours consistently produces representations where the object and its class remain clear, highlighting its robustness and effectiveness.

%In Fig. \ref{fig:fig_class_neurons}, we illustrate example class neuron visualizations. Unlike prior methods, which often produce repetitive and unrecognizable patterns, especially with large-scale models such as ResNet-50, our approach achieves more interpretable visualizations by aligning generated images with meaningful distributions. While our method generally offers clearer, conceptually relevant images, there are still some cases where visualization quality suffers, particularly in images with complex scenes, such as the \textit{vacuum cleaner} and \textit{ambulance}, where the concept is less distinct. These negative examples highlight areas for further refinement of our framework. We provide more qualitative results in App..

%In Fig. \ref{fig:fig_inner_neurons}, we expand our qualitative results for intermediate neuron visualizations. The neurons are selected based on their relevance to a particular class neuron, identified using LRP. This approach provides a deeper insight into the neuron behavior and enables more effective comparison across methods. We provide more qualitative results in App. for randomly selected neurons from the penultimate layer.

\subsubsection{Analysis of the Embedding Space}

\begin{figure}
    \centering
    \centerline{\includegraphics[width=\linewidth]
    {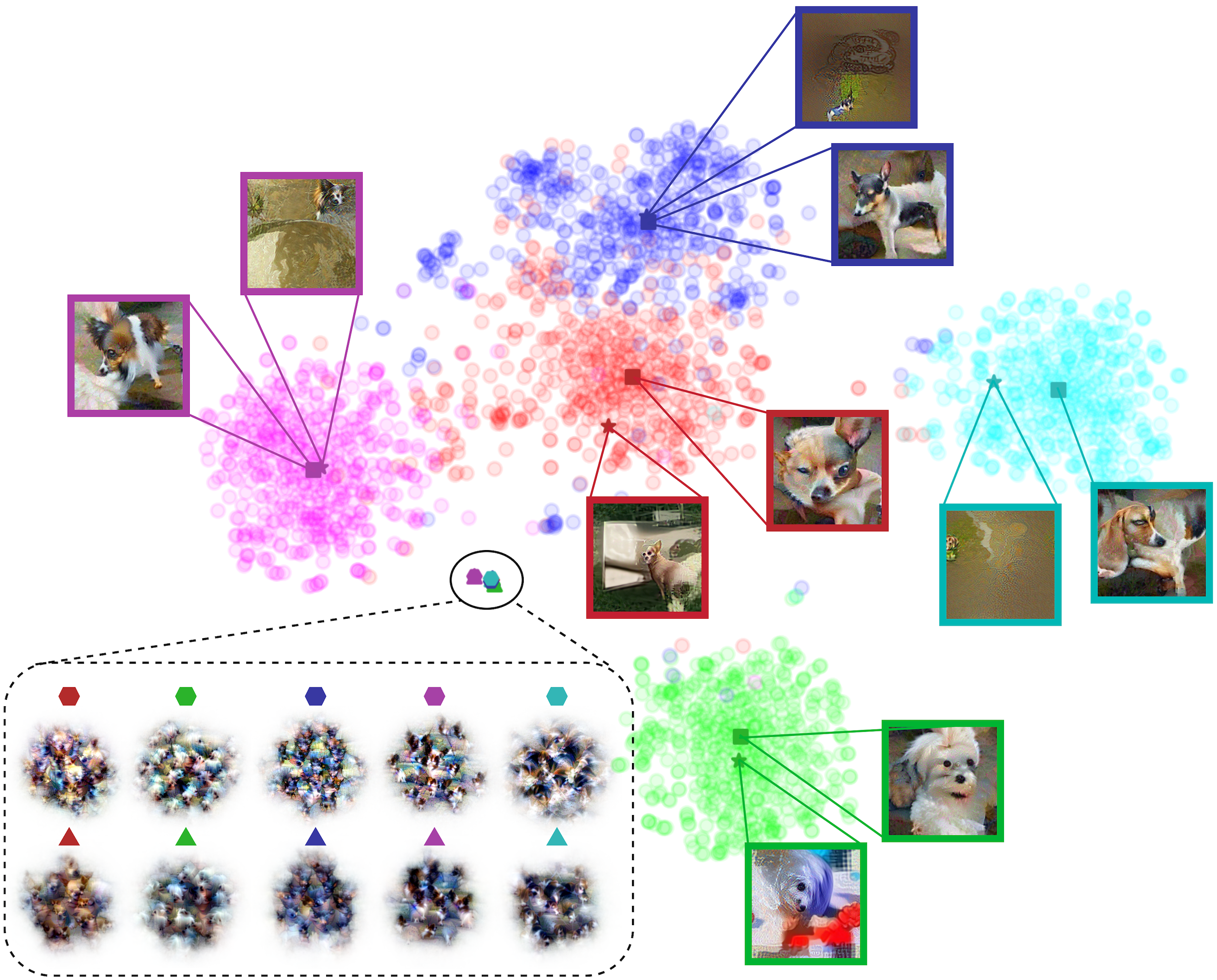}}
    \caption{\textit{t-SNE projection of embedding.} We show a low-dimensional tSNE embedding of the features at the penultimate layer for five dog breeds indicated by color. Transparent circles are original training images and FVs are indicated by symbols:  \\
    $\blacksquare$: VITAL, $\blacktriangle $: MACO, $\hexagonblack$: Fourier, $\bigstar$: DeepInv.}
    \label{fig:fig_tsne}

\end{figure}

To understand how realistic and distinct the visualizations are, we compare FVs of five different dog classes to original images in terms of their penultimate layer embedding.
We show a 2D low-dimensional representation of the embeddings of original and generated images using tSNE~\cite{tSNE} in Fig.~\ref{fig:fig_tsne}.
We find that the real images form five clusters neatly separating the classes. 
We further observe that fourier-based approaches with their repetitive, often unrecognizable patterns are collapsed on a single point, away from any real images.
While DeepInversion does fall into the correct clusters, it is hard to recognize a dog in the generated images.
\ours is the only method that consistently falls into the center of the cluster, providing distinct dog features we can recognize as known dog breeds. We extend our results for other architectures in Appendix \cref{subsec:supp_architectures}.

%We further analyze the quality of the generated images through visualizations of the embedding space using t-SNE \cite{tSNE} to compute a 2D low-dimensional representation. This way, we can assess how well the generated images capture the intrinsic representations of real images. It helps us clarify whether what the model sees, expects, and learns aligns coherently with the generated images. In Fig., we give an example using classes of five different dog breeds. From the results, we can see VITAL's ability to act like cluster centers whereas MACO, Fourier, and CBR collapsed into a single point, away from the clusters of real images. Considering the results of both VITAL and DeepInv, we can conclude that using statistics to model the data is essential to have a more realistic representation in the embedding space. We provide more results for different classes, including visualizations using multidimensional scaling (MDS) \cite{MDS} in the App..

\subsubsection{Small Circuits}

\begin{figure*}[t!]
    \centering
    \centerline{\includegraphics[width=\linewidth]
    {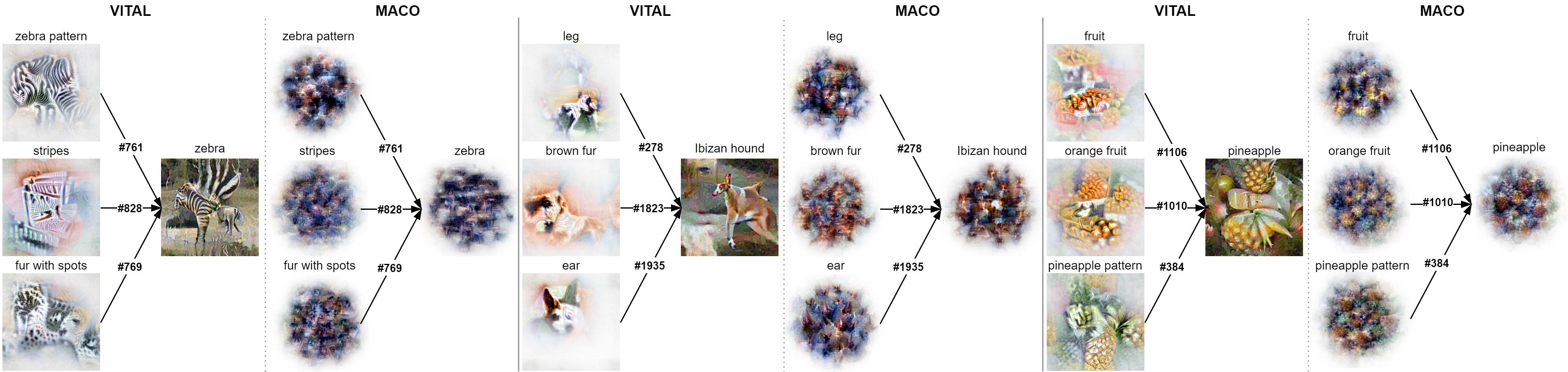}}
    \caption{\textit{Visualizations of small "circuits".} We provide visualizations of small "circuits" for three interesting classes of ImageNet. For each class, we found the three most relevant neurons in the penultimate layer based on LRP relevance scores and indicate the neuron ID on the arrow. For each neuron, we label its likely meaning and provide FVs by MACO and \ours. The Zebra class shows high dependence on the characteristic pattern of the fur, the dog class a dependence on specific color patterns to be able to distinguish the abundant different dog classes in ImageNet, and Pineapple shows signs of overfitting with an association with collections of fruits.
        % \bernt{I am wondering if we should disentangle the visualization here between MACO and VITAL - imo we should aim for one figure each for MACO and VITAL side by side - and then still for three cases as shown. That shoudl make the difference between MACO and VITAL farm more striking?}
        }
    \label{fig:fig_inner_neurons}

\end{figure*}

% With our approach, it is possible to investigate what inner neurons in a network encode.
% By extracting the top-3 channel in the last hidden layer most relevant to a specific class ranked by relevance through LRP, we can visualize small mechanistic \textit{circuits} that show how specific patterns drive a class decision (see Fig.~\ref{fig:fig_inner_neurons}).
Our method enables interpreting what inner neurons encode. By selecting the top-3 channels in the last hidden layer most relevant to a class (via LRP), we can visualize local \textit{circuits} that reveal class-specific patterns (Fig.~\ref{fig:fig_inner_neurons}). 
% We note that except for DeepInversion, all approaches can be used to visualize inner neurons. For illustrative purposes, we selected MACO as it is the best-performing existing approach.
While most methods, except DeepInversion, support inner neuron visualization, we compare with MACO as the strongest baseline.
For example, we see that the most relevant, distinguishing features for a zebra classification are the stripes and texture of their fur. For dogs, there are leg, ear, and specifically colored fur detectors evident in the model. Such specific coloring is necessary to distinguish the plenty of dog breeds available as classes in ImageNet. We also observe signs of overfitting, where the pineapple class is associated with neurons that detect groups of fruits, many of the training images show such a picture.
% While MACO also yields inner neuron visualizations, for the animal classes they are often inherently hard to understand, which we will also see in quantitative terms next.
MACO yields less interpretable results, particularly for animals, as confirmed in the next section.

\subsection{Quantitative Results}
\label{subsec:exp_quant}

For quantitative evaluation, we consider the classification accuracy and the FID score~\cite{Seitzer2020FID} on the respected model, as well as CLIP zero-shot prediction performance~\cite{CLIP} in \cref{tab:combined_new} (see Appendix \cref{tab:supp_all} for the extended version). We also provide quantitative results for the intermediate neurons with the metrics proposed by \cite{kopf2024cosy} in Appendix \cref{subsec:supp_performance_inner}.
%We exclude methods that utilize auxiliary architectures such as GANs \cite{FV_GAN} or autoencoders \cite{FV_Autoencoder,Wang2022TraditionalCN} to ensure a fair comparison and to promote transparency as it is unclear whether visualized features are from network under study or the generating network ("hallucination"). 

\myparagraph{Accuracy.} 
% Visualizing each class neuron and then predicting the class of the visualization, we measure how accurate of a description of the target these visualizations are (see Tab.~\ref{tab:combined_new}).
To assess how well visualizations describe their target, we classify each class neuron’s visualization (see Tab.~\ref{tab:combined_new}). 
% While existing work optimizes for class logits, perhaps surprisingly, the achieved accuracy for all methods except DeepInversion is low.
Despite explicitly optimizing class logits, most existing methods perform poorly.
Only \ours and DeepInversion are able to produce accurate descriptions, achieving virtually 100\% accuracy. Investigating further, we observe that other methods including MACO seem to produce irrelevant features that mislead the model, producing predictions unrelated to the original class. We give two examples of prediction in Appendix \cref{subsec:supp_classification}.

%In \cref{tab:accuracy}, we show the top-2 softmax  scores of synthetic images, as well as mean across 100 ImageNet train images of that class, across 2 randomly selected classes. Top-2 performing classes enables us to evaluate the robustness of the visualizations in capturing the responses of real data as well as ensuring the network to confidently distinguish between multiple plausible classes for a given image. While we previously denoted the qualitative results as more interpretable, the classification scores also show that they are more aligned with class-relevant features that the network has learned as well. Also, in \cref{tab:combined}, we show the top-1 accuracy of the images for all classes. It should be noted that DeepInv is trained with a classification loss and therefore is expected to be the best as observed in \cref{tab:accuracy,tab:combined}.

\myparagraph{FID Score.} 
% For all architectures, to quantify the similarity between the feature distributions of visualizations to that of natural images within the same object category in \cref{tab:combined_new}, we use FID scores~\cite{Seitzer2020FID} computed on features extracted at the penultimate layer of ResNet50.
For all architectures, to quantify visualization realism, we compute FID scores~\cite{Seitzer2020FID} using features from ResNet50’s penultimate layer (\cref{tab:combined_new}). 
% While this metric is aligned with our optimization, it is surprising to see the extent of the difference to most other methods, showing over a magnitude better score. 
While aligned with our objective, the performance gap to other methods is substantial, with \ours achieving over an order of magnitude better scores.
% Similar to \ours, DeepInversion also considers feature statistics, namely batch normalization statistics. While in the optimal case reaching the best performance, the choice of batch size can have a high impact on visualization quality and the FID score.
DeepInversion also leverages feature statistics but is highly sensitive to batch size, which impacts both FID and visual quality.

\myparagraph{CLIP Zero-shot Prediction.} 
% So far, previous metrics were based on the model for which visualizations were generated.
So far, previous metrics relied on the model used for visualization.
% To understand how a different model perceives the visual features presented in a FV, we consider a pretrained CLIP ViT-B/32 model and query it with templates to generate class labels for the FVs (see Appendix for more details).
To assess general perceptual quality, we use a pretrained CLIP ViT-B/32 model to classify FVs via template-based prompts (see Appendix \cref{subsec:supp_clip} for more details). 
% Based on these answers, we compute a zero-shot prediction score for all classes and each method.
We report zero-shot scores in Tab.~\ref{tab:combined_new}, including original (correctly classified) ImageNet images for reference. 
% We provide scores including those of original (correctly classified) ImageNet images in Tab.~\ref{tab:combined_new}. 
As before, we see that DeepInversion outperforms existing work, with standard baselines having overall poor performance even when considering Top-5 accuracy.
Notably, \ours outperforms all others by a wide margin, approaching real image accuracy.
% Furthermore, evaluated on this \textit{separate} model, \ours surpasses the performance of all existing work by a wide margin, coming close to that of real data images.

\begin{figure}
    \centering
    \begin{subfigure}{0.98\linewidth}
    \centering
    \includegraphics[width=.81\linewidth]{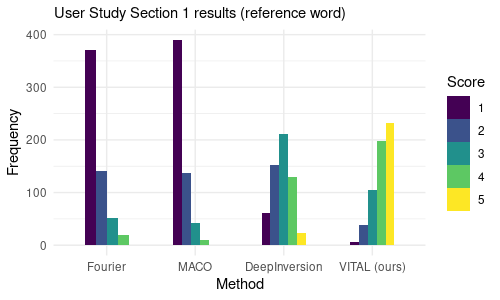}
    \caption{Class visualization with given class label.}
    \end{subfigure}
    \newline
    \begin{subfigure}{0.98\linewidth}
    \centering
    \includegraphics[width=.81\linewidth]{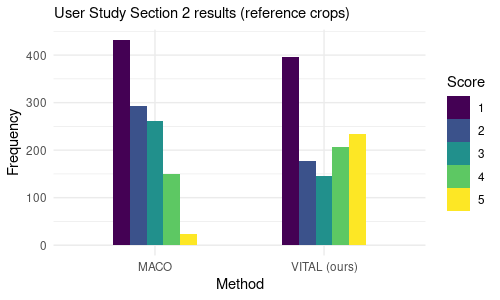}
    \caption{Inner neuron visualization with given activating images.}
    \end{subfigure}
    \newline
    \begin{subfigure}{0.98\linewidth}
    \centering
    \includegraphics[width=.81\linewidth]{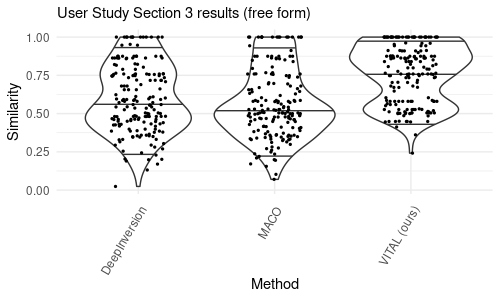}
    \caption{Class visualization with free-form labeling.}
    \end{subfigure}%
    \caption{\textit{Human Study results.} We provide summary statistics on the scores for the different methods obtained for the three-part human ($n=58$) evaluation study. In (a), participants were given class visualization and corresponding class name for $10$ random classes and scored the visualizations on a scale of 1(worst) to 5(best). In (b) the task was to evaluate the closeness of FVs for an inner neuron to the image crops that it most activates on (same scoring). DeepInversion does not generate visualizations for inner neurons. In (c), participants were asked to label a given FV. We evaluated similarity of their label to the (hidden) ground truth based on Universal Sentence Encoder embeddings. We provide violin plots with median and 5\% and 95\% quantiles of the achieved similarity.}
    \label{fig:humanstudy}
\end{figure}

\subsection{Human Interpretability Study}

The primary objective of FVs is to make neural networks more understandable \textit{for human users}. To address this goal, we conducted a user study (58 participants) to evaluate the interpretability of a FV from the perspective of a human. Specifically, participants (1) given a single word, evaluated how well a FV reflects the provided word, (2) evaluated how well the FVs for an \textit{inner neuron} reflect the provided reference images (highly activating on the target neuron), and (3) should describe a given generated image with a word or short description. These present settings of increasing difficulty and realism. While (1) is a typical evaluation, it also comes with the largest implicit bias, as the class label already gives away what we want to visualize. In (2), by showing reference images, we encourage a \textit{visual match} between FV and real image. We compare with MACO, as it is the best performing approach to visualize inner neurons. In (3), we model a real-world scenario, i.e., without any implicit bias, the user is supposed to understand what the network encodes.
We provided a simple scoring system from 1(worst) to 5(best) for parts (1) and (2), and evaluated how well the textual description in (3) fits the class using closeness in the embedding space of a text encoder~\cite{USE}.

We provide an overview of the results of our study in Fig.~\ref{fig:humanstudy}.
We observe that our method performs favorable across all three tasks compared to other methods, showing an order of magnitude more high-scoring visualizations (scores 4-5) in both (1) and (2), and showing much higher similarity to the correct class on task (3).
In particular, for this challenging task that provides a zero-shot estimate of the interpretability of FVs, the 5\% quantile, i.e., the "worse"-performing visualizations, are still close to a similarity of $.5$ of \ours. Other methods show only $0.25$ similarity for the same quantile. The same stark difference of more than $0.2$ in median similarity score shows that \ours provides more human-understandable visualizations of what neurons encode. We provide more fine-grained analysis, population statistics, and details in Appendix \cref{subsec:supp_user_study}.
We further conducted a human user study following the protocol of \cite{zimmermann2021causal, Fel2023}, which is  a simple option-based evaluation similar to setting (2), where VITAL performs best (Appendix \cref{subsec:supp_user_study}).

% \subsection{Ablation Study}
% \label{subsec:ablations}

% We performed an analysis of how different building blocks of a network affect the final visualization. As illustrated in Fig. \ref{fig:fig_ablation_layer}, when we match the feature distribution of the coarser layers, the resulting visualization primarily captures low-level information such as colors and textures, which we refer to as the style information. As we go deeper into the network, the visualizations progressively incorporate more contextual information such as the shape or the structure of an object at the cost of increased high-frequency noise. Accordingly, we observed that we can achieve a more realistic and proper visualization result by employing all the building blocks of our model that enable us to transfer the style into the context.

% \myparagraph{Effect of the Regularization Losses.} 
% I will include this here based on our space.

% \subsection{Concept visualizations and polysemanticity}

% TODO: show some kind of "application" showing that it is complementing "mechanistic circuits". More of a salespitch here, just apply it on CRAFT or so. Would fit well into the overall story and into the Mechanistic Interpretability landscape. Fig. 7 etc.

\section{Discussion}
\label{sec:discussions}

% Our experiments show that \ours yields \textit{better understandable} visualizations, more accurately capturing the distinct information encoded by a neuron, outperforming its most recent competitor MACO~\cite{Fel2023} in terms of classification accuracy for the generated visualizations on the target model.
% While DeepInversion shows similar performance on this metric, this comes to little surprise, as their approach explicitly maximizes the logit class score of that (same) model.
% When asking a \textit{different} neural network, such as a CLIP model, to label the images, however, our approach outperforms all competitors by a wide margin.
% Similarly, two human user studies we conducted provide evidence that visualizations of \ours are more understandable than those of competitors in describing given labels or images.
Our experiments show that \ours produces more understandable visualizations that better capture the unique information encoded by each neuron, outperforming MACO~\cite{Fel2023}, the most recent method in classification accuracy on the target model. While DeepInversion performs similarly, this is expected since it directly optimizes the model’s logit score. When evaluated by a different model such as CLIP, \ours surpasses all baselines by a wide margin. Two human user studies further confirm that \ours yields more interpretable visualizations in describing given labels or images.

\myparagraph{Realism.} 
In principle, our visualizations, as well as those of other methods, do not reach photo-realism. \ours visualizations sometimes resemble the style of paintings of \'Edouard Manet, which, however, does not limit their interpretability: Similar to a painting, it is \textit{understandable} what is expressed, which is also supported by the user studies.
Yet, it would be interesting to study how to reach photo-realism in the future.
Generative models such as diffusion offer a promising direction, though further research is needed to prevent the introduction of model-specific artifacts that reduce faithfulness to the target model.
% While generative approaches such as diffusion models is an interesting avenue, it requires further research on how to avoid generative-model specific features in the generated image, as these erode the faithfulness of the visualizations.
% While our visualizations, like those of other methods, do not achieve photorealism, \ours sometimes resemble the style of Édouard Manet’s paintings. This does not limit their interpretability; like a painting, the content remains understandable, as confirmed by our user studies. Achieving photorealism remains an open challenge. Generative models such as diffusion offer a promising direction, though further research is needed to prevent the introduction of model-specific artifacts that reduce faithfulness to the target model.

\myparagraph{Evaluation.}
% For our evaluation, we considered a diverse set of metrics to evaluate our framework against the state-of-the-art as comprehensively as possible, leveraging what is common in the literature and adding further metrics, such as through CLIP prompting.
We evaluate our framework using a broad set of metrics, combining standard benchmarks with additional measures like CLIP prompting.
Unfortunately, in feature visualization, more broadly in Explainability research, exhaustive benchmarks and clearly defined metrics are actively discussed, but currently still amiss.
Through the additional human user studies, we complement the metrics with an evaluation of our original goal of \textit{more understandable feature visualizations}.
% As with all human user studies, its design can always be debated. 
% More refined study settings and including larger cohorts would perhaps make for an interesting study on its own.
As with any user study, the design can be debated, and more refined setups with larger cohorts offer a promising direction for future work.

\myparagraph{Distribution Matching.} Standard approaches to distribution matching include matching statistics, such as mean and variances, or considering prior distributions \cite{AdaIN,UniStyle}, but these oversimplify the complexity of real images. 
% We observed a diverse set of activation distributions that are hard to capture with a (simple) distribution prior or first or second moment statistics \cite{SM}. Hence, directly optimizing an alignment between the \textit{empirical} distributions showed the most promise, with sort matching~\cite{rolland2000} providing an efficient approach that we can scale easily to multiple layers of modern networks.
Activation distributions are often too diverse for such approximations~\cite{SM}. Hence, directly optimizing an alignment between the \textit{empirical} distributions showed the most promise, with sort matching~\cite{rolland2000} as an efficient approach that easily scales to modern networks, but our approach is agnostic to the type of distribution matching algorithm.

\myparagraph{Future work.} %Both human user studies showed that generated images, regardless of method, are not consistently scored as perfectly reflecting the given class or image.
\ours provides a substantial improvement over existing work and we anticipate it of great use in downstream applications, such as understanding networks in the medical domain, studying what neurons encode after knowledge transfer, or how pruning affects the representations.
Future directions include automatic selection of neurons or neuron sets for circuit visualization, for example using CRP~\cite{CRP}, integration of ViT-specific visualization via AttnLRP~\cite{AttnLRP}, and extending our analysis to CLIP-based ViT models to better understand multimodal representations.

\section{Conclusion}
\label{sec:conclusion}

% Feature visualization (FV) is crucial for understanding neural network encoding, yet existing methods often produce repetitive patterns, irrelevant features, or artifacts. We reframe FV by optimizing for feature distribution similarity between generated and real images. We further suggest \textit{incorporating relevance scores} to accurately capture the information that is encoded in an intermediate neuron, helping to avoid irrelevant features. Through diverse metrics and two human user studies, we show that \ours provides clearer, more interpretable visualizations for both humans and machines while better capturing the information encoded in a target neuron. We anticipate \ours to aid safety-critical domains like medicine and complement Mechanistic Interpretability by assigning meaning to mechanistic circuits.
Feature visualization is key to understanding neural representations, but existing methods often yield artifacts or irrelevant patterns. We propose a new approach that aligns the feature distributions of generated and real images, and incorporates relevance scores to better capture the true information encoded in intermediate neurons. Across multiple metrics and two user studies, \ours produces clearer and more interpretable visualizations for both humans and models. We envision \ours supporting safety-critical fields such as medicine and complementing Mechanistic Interpretability by attributing meaning to neural circuits.

% Feature visualization (FV) are an essential tool for understanding the internal encoding of neural networks.
% Existing works on FV struggle with accurately capturing the information encoded by a target neuron, showing repetitive patterns, spurious features, or artifacts.
% We propose to reframe the problem of FV, optimizing for a \textit{similarity of feature distributions} between generated and real images.
% We further suggest \textit{incorporating relevance scores} to accurately capture the information that is encoded in an intermediate neuron, helping to avoid spurious features.
% Considering a diverse set of metrics and a human user study, we show that in contrast to existing work, our method \ours provides \textit{understandable} visualizations to both human and machine better capturing the information encoded in a target neuron.
% We anticipate \ours to help in safety-critical down-stream domains, such as medicine, where understanding the network is key, but also to serve as a complementary tool for Mechanistic Interpretability, giving \textit{meaning} to mechanistic circuits.
{
    \small
    \bibliographystyle{ieeenat_fullname}
    \bibliography{main}
}
% \clearpage
% \maketitlesupplementary
% \renewcommand{\thesection}{\Alph{section}} 
% \setcounter{section}{0} % Reset section counter to start from A
% \input{sec/supp_Method}
% \input{sec/supp_Results}

\clearpage
\appendix  % Correctly signals start of supplementary material
\maketitlesupplementary
\renewcommand{\thesection}{\Alph{section}} 
\setcounter{section}{0} % Reset section counter to start from A

% Ensure cleveref understands the new numbering
\crefname{section}{Appendix}{Appendices}
\Crefname{section}{Appendix}{Appendices}

\section{Method}
\label{sec:supp_method}

\begin{figure*}
    \centering
    \centerline{\includegraphics[width=.95\linewidth]
    {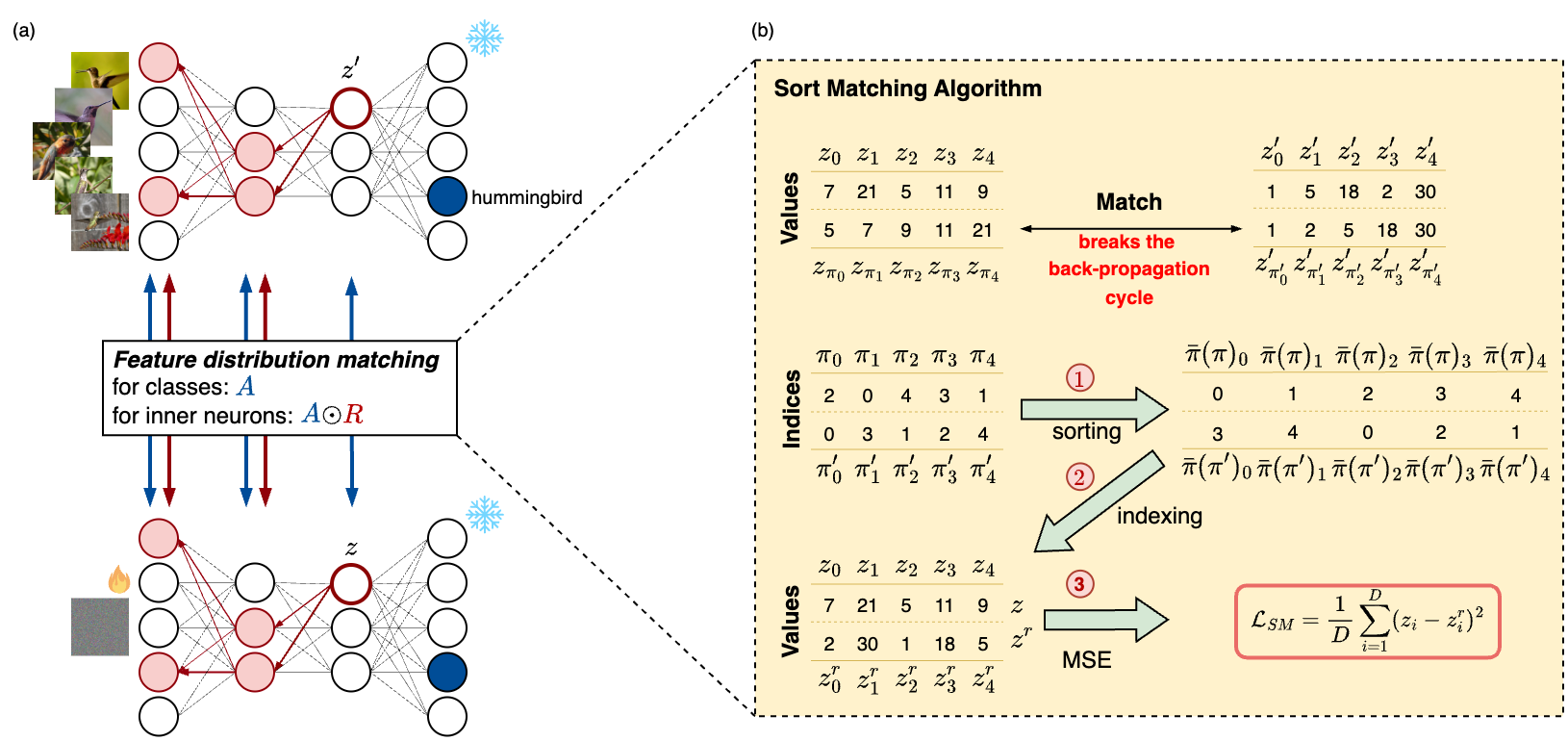}}
    \caption{\textit{Overview of VITAL framework.} Our \ours framework mainly consists of two main components. In part \textbf{(a)}, we utilize a pretrained \textit{frozen} model  (\textcolor{cyan}{\ding{100}}) to generate visualizations. For the given type of visualization-class neurons or intermediate neurons-we first select $N$ reference images from the ImageNet training dataset, then we optimize a randomly initialized image (\includegraphics[width=0.2cm]{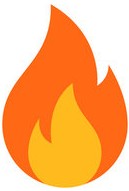}) through our feature distribution matching approach, which is applied across multiple layers of the model. For feature distribution matching, we utilize the feature activations \textcolor{activation_color}{$A$} and feature relevance (\textcolor{arrow_color}{arrows}) \textcolor{activation_color}{$A$}$\odot$\textcolor{arrow_color}{$R$} for class neurons and intermediate neurons, respectively. We perform the feature distribution matching using \textbf{(b)} our sort matching loss, where we compute the difference between the feature distributions of $z$ (from synthetic data) and $z'$ (from reference data). We achieve this by first sorting the values and obtaining the sorted indices as $\pi$ and $\pi'$. Considering that sorting is a discrete operation and we want to allow back-propagation to optimize $z$, we obtain a reverse mapping $\Bar{\pi}(\cdot)$ by  \textcolor{arrow_color}{\ding{172}} sorting the indices $\pi'$ to \textcolor{arrow_color}{\ding{173}} re-index $z'$ to $z^r$, which allows $z$ to be unchanged. Thus, we were able to \textcolor{arrow_color}{\ding{174}} match $z$ and $z^r$ through minimizing the MSE loss. The proposed SM loss can be used in a plug-and-play manner without introducing any parameters, as summarized in Alg. \ref{alg:algo1}.
    }
    \label{fig:fig_method_supp}

\end{figure*}

Here, we provide detailed information about the \ours framework, the sort-matching procedure, as well as further implementation considerations for all tested methods. In \cref{fig:fig_method_supp}, we give an overview of \ours, with an example computation of the sort matching (SM) algorithm and its corresponding pseudo-code provided in Alg. \ref{alg:algo1}.

\subsection{Implementation Details}
\label{subsec:supp_imp}

For the experiments, we use a publicly available pretrained models (ResNet50 \cite{He2015DeepRL}, DenseNet121 \cite{densenet}, ConvNeXt-base \cite{convnext}, ViT-L-16 \cite{vit}, ViT-L-32 \cite{vit}) from the PyTorch \cite{PyTorch} library. We report feature visualizations (FVs) of all methods across three different random seeds for each category of the ImageNet dataset \cite{deng2009imagenet}. 
In detail,

\begin{itemize}
    \item for \textbf{VITAL}, we synthesize a single image with resolution $224 \times 224$ and apply jittering at each optimization step to promote robustness. We set the number of real images in our reference dataset $\mathcal{X}_{ref}$ for the feature distribution matching as $N=50$. For the polysemanticity experiments further below, we similarly set $N=50$ and consider $1000$ patches for $k$-Means. For optimizing the feature visualization, we use Adam with a learning rate of $1.0$. For intermediate neuron visualizations, we select the patch size as 64 and set the scales $\alpha_{\text{TV}}=\alpha_{\ell_2}=3\times 10^{-6}$, $\lambda=1$. We provide ablations for the effects of $\alpha_{\text{TV}}, \alpha_{\ell_2}$ in Sec.~\ref{subsec:supp_ablation_hyp}. After analyzing the effect of each network component of ResNet50 on our SM loss in Sec.~\ref{subsec:supp_ablation_layers}, we decided to utilize all the network components. Specifically, For ResNet50, for class neurons, the loss weight for each block is set to $1.0$ whereas for intermediate neurons, we reduce the contribution of block1 to be $0.1$. For DenseNet121 with class neurons, the loss weight for each block is set to $1.0$ except the final block, which is set to $100.0$. For ConvNeXt-base with class neurons, the loss weight for each block is set to $1.0$ except the first block, which is set to $10.0$. For ViT-L-16 with class neurons, for the selected 5 blocks that includes the projection layer and selected encoder layers, the loss weight for each block is set to $1.0$. Finally, for ViT-L-32 with class neurons, for the selected 5 blocks that includes the projection layer and selected encoder layers, the loss weight for each block is set to $1.0$ except the final block, which is set to $0.1$.  
    For the experiments involving the visualization of class neurons using LRP with ResNet50, we additionally utilized auxiliary regularization with parameters $\alpha_{\text{TV}}=\alpha_{\ell_2}=0.00001$. 
    \item for \textbf{DeepInversion} \cite{yin2020dreaming}, we synthesize a batch of images with resolution $224 \times 224$ and apply jittering at each optimization step to promote robustness.  We adapted the parameters from their official GitHub implementation \cite{DeepInv_github}. In detail, we use Adam for optimization with a learning rate of $0.05$, and set the scales of the auxiliary regularization as $\alpha_{\text{TV}}=0.0001$, $\alpha_{\ell_2}=0.00001$, $\lambda=1$. For a fair comparison, we do not apply the teacher-student guidance that was proposed in Adaptive DeepInversion. 
    \item for \textbf{MACO} \cite{Fel2023}, we synthesize both regular ($224 \times 224$) and high resolution ($1024 \times 1024$) images offered by MACO and we found that higher resolution visualizations were more human readable. Yet, as shown in the quantitative experiments, this effect seemed more like a subjective, qualitative finding and did not carry over to CLIP Zero-shot prediction scores, classification scores, or FID scores. As suggested by \citet{Fel2023}, for transformations, we first add uniform noise $\delta \sim \mathcal{U}([-0.1, 0.1])^{W \times H}$ and augment the data at each iteration with crops of the input image that are resized to ($224 \times 224$), in which the crop size drawn from the normal distribution $\mathcal{N}(0.25, 0.1)$. For optimization, we use the Adam optimizer with a learning rate of $1.0$. 
    \item for \textbf{Fourier} \cite{olah2017feature}, we use the same settings as for MACO, only the initialization of the generated image is changed to regular Fourier initialization, i.e., without fixed magnitude.
    \item for \textbf{PII} \cite{pmlr-v162-ghiasi22a}, we use the published implementation, including the provided batch-size settings for all models except ConvNext-base (not implemented) with image resolution of $224 \times 224$.
\end{itemize}

% \begin{algorithm}
%     \caption{SM Loss}
%     \label{alg:algo1}
%     \SetAlgoLined
%     \begin{algorithmic}[1]
%         \State \textbf{Input:} Two numbers $a$, $b$. 
%         \State $-, \text{IndexX} = \text{torch.sort}(x)$ \hfill \# Sort $x$ values
%         \State $\text{SortedY}, - = \text{torch.sort}(y)$ \hfill \# Sort $y$ values
%         \State $\text{InverseIndex} = \text{IndexX.argsort}(-1)$ 
%         \State \textbf{return} $x + \text{SortedY.gather}(-1, \text{InverseIndex}) - x.\text{detach}()$
%     \end{algorithmic}
% \end{algorithm}

% Redefine \Require to print "Input"
\renewcommand{\algorithmicrequire}{\textbf{Input:}}

\begin{algorithm}
    \caption{SM Loss for layer-$l$}\label{alg:algo1}
    \begin{algorithmic}
    \Require $f_l(x) \subseteq \mathbb{R}^{1 \times C \times HW} $, $f_l(y) \subseteq \mathbb{R}^{N \times C \times HW} $
    \State $-, \text{IndexX} = \text{torch.sort}(f_l(x), \text{dim}=2)$ \hfill 
    \State $\text{SortedY}, - = \text{torch.sort}(f_l(y), \text{dim}=2)$ \hfill 
    \State $\text{SortedY} = \text{torch.mean}(\text{SortedY}, \text{dim}=0)$
    \State $\text{InverseIndex} = \text{IndexX.argsort}(-1)$ 
    \State $g_l(y)=\text{SortedY.gather}(-1, \text{InverseIndex})$
    \State \textbf{return} $\mathcal{L}_{MSE}=\text{torch.mean}((f_l(x)-g_l(y))^2)$
    \end{algorithmic}
\end{algorithm}

\subsection{CLIP Zero-shot Prediction}
\label{subsec:supp_clip}

To evaluate the FVs based on how "understandable" they are in terms of the target class they aim to visualize, we considered a pretrained CLIP model.
The CLIP space \cite{CLIP} is an effective method for quantifying visualization methods as it bridges the gap between image and text, thus offering a powerful measure of how well a visualization aligns with an intended concept. We perform this experiment to understand how a \textit{different} model \textit{perceives} the visual features presented in a FV. Specifically, we load a pretrained CLIP ViT-B/32 model and expand the ImageNet labels into descriptive textual prompts (80 templates) \cite{CLIP_zero_shot}, such as "a photo of a \{label\}" or "a picture of a \{label\}". The main goal is to explore how different descriptions of the same class affect the model's predictions by varying the phrasing of these prompts. We divide the calculation of the zero-shot prediction scores into two steps. In the first step, we first associate each class label with the textual prompts and for each prompt, we compute its embedding using CLIP's text encoder. These embeddings are then averaged to obtain a single representative embedding for each class, which we refer to as their corresponding "zero-shot weight". Each class is hence represented by a robust and generalized text embedding. In the second step, we compute the embeddings of the input images, including our FVs, into CLIP's shared feature space using its respective encoders. Then, we compute the cosine similarities between the zero-shot weights of each class and the image embeddings to identify the best-matching class for each image. The performance is evaluated by measuring the zero-shot classification accuracy on  the original (correctly classified) ImageNet images as well as the images of different FV methods.

\subsection{Selection of Reference Images}
\label{subsec:supp_selection}
We define our image set as the entire ImageNet training data. Considering the computational needs, a subset of the training data could also be used. After defining our image set, we divide our selection of $\mathcal{X}_{ref}$ for the two sub-problems, (1) class visualizations and (2) intermediate neuron visualizations. For (1), we select $N$ random images for feature distribution matching. In (2), we first select sub-regions to identify specific concepts that neurons respond to within a localized context. Similar to CRAFT \cite{CRAFT}, we first crop and resize the training images into patches to obtain an auxiliary dataset. Then, we obtain the activations for each patch, keeping the top-$N$ patches while eliminating the patches that are coming from the same original image. In the case where the model is a CNN, the score for each patch is formulated as the global average pooling of activations across their spatial dimension.

\begin{figure}
    \centering
    \centerline{\includegraphics[width=\linewidth]
    {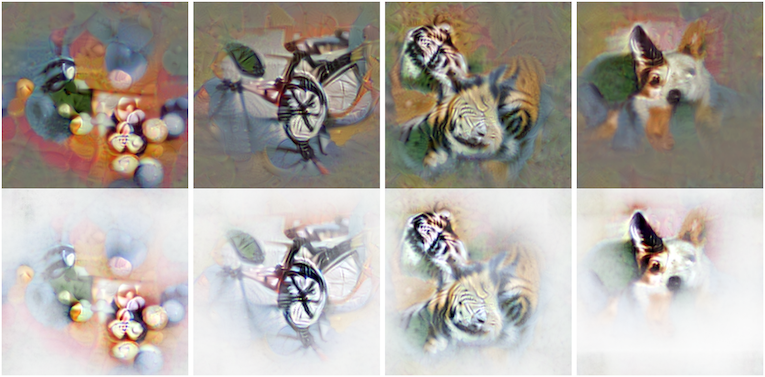}}
    \caption{The effect of the transparency map in the \ours framework. The \textbf{first row} represents the visualization \textit{without} a transparency map and the \textbf{second row} represents the visualization \textit{with} a transparency map.}
    \label{fig:fig_transparency}

\end{figure}

\begin{figure}
    \centering
    \centerline{\includegraphics[width=\linewidth]
    {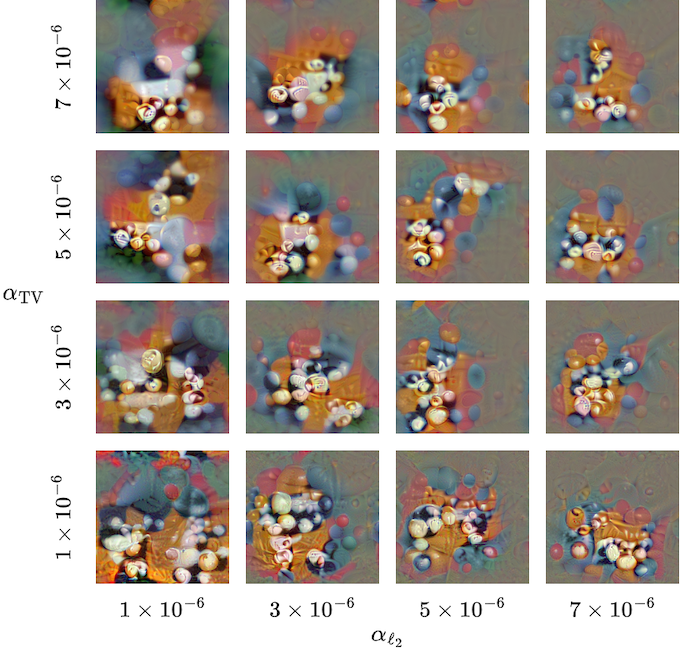}}
    \caption{The effects of the parameters total variation $\alpha_{\text{TV}}$ and 
$\ell_2$ norm $\alpha_{\ell_2}$ on the final visualization in the \ours framework.}
    \label{fig:fig_hyperparameter}

\end{figure}

\begin{figure}
    \centering
    \centerline{\includegraphics[width=\linewidth]
    {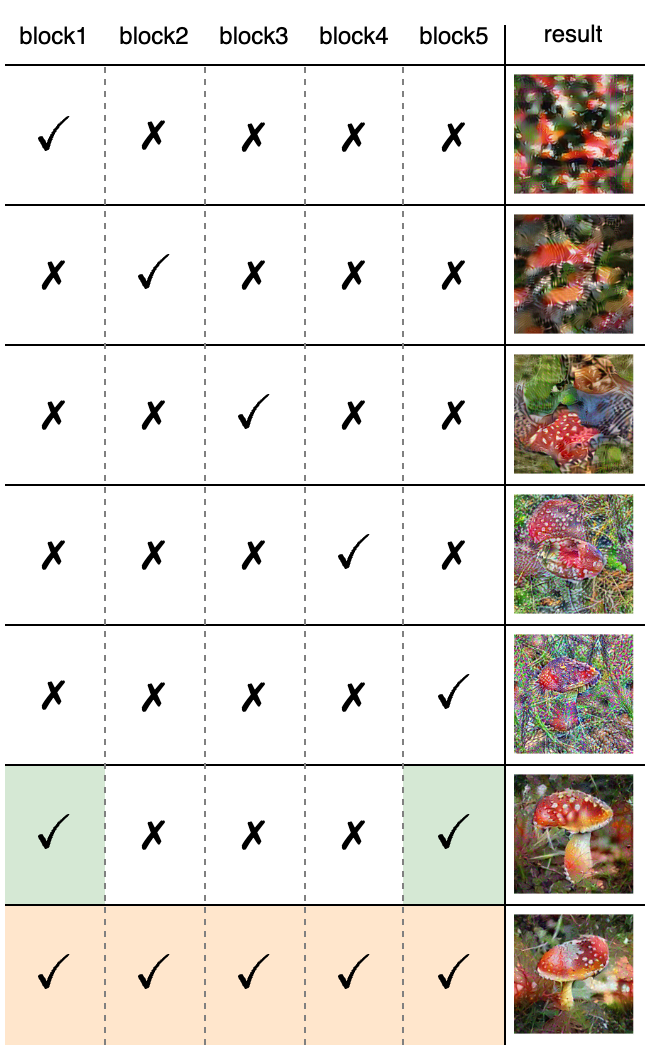}}
    \caption{The effects of individual model components and their combination on the final visualization in the \ours framework. Each block in the columns refers to a network component of ResNet50 (e.g., block1:conv1, block5:layer4).}
    \label{fig:fig_ablation_layer}

\end{figure}
\begin{table*}[t]
    \centering
    \scriptsize
    \begin{tabular}{l cccccccc}
        \toprule
         & \multirow{2}{*}{Method} & \multirow{2}{*}{Setup} &Acc. &  \multicolumn{2}{c}{FID ($\downarrow$)}& \multicolumn{2}{c}{Zero-Shot Prediction}   \\ 
         \cmidrule(lr){5-6} \cmidrule(lr){7-8} 
         & & & Top1 ($\uparrow$) & RN50 & Arch.  & Top1 ($\uparrow$)  & Top5 ($\uparrow$) \\
         \midrule
         \multirow{14}{*}{\rotatebox{90}{ResNet50}} &\textit{ImageNet} &  & - & -& -&  69.11 & 92.23 \\
         \cmidrule{2-9}
         & \multirow{2}{*}{MACO} & r: 224 & 29.43 & 360.74 & 360.74 & 12.87 & 29.73 \\
         && r: 1024 & 2.10 & 494.57 & 494.57 & 1.60 & 5.67   \\
         \cdashline{2-9}\noalign{\vskip 1.0ex}
         &\multirow{2}{*}{Fourier} & r: 224 & 21.30 & 422.44 & 422.44 & 6.73 & 18.27  \\
         && r: 1024 & 3.43 & 430.58 & 430.58 & 0.97 & 3.57  \\
         \cdashline{2-9}\noalign{\vskip 1.0ex}
         % &\multirow{2}{*}{CBR} & r: 224 & 0.50 & 545.50 & & 545.50 & 0.20 & 1.17  \\
         % && r: 1024 & 0.33 & 522.89 & & 522.89 & 0.20 & 1.07  \\
         % \cdashline{2-9}\noalign{\vskip 1.0ex}
         &\multirow{2}{*}{DeepInversion}& bs: 64 & \textbf{100.00} & \textbf{35.76} & \textbf{35.76} & \underline{29.90} & \underline{55.20} \\
         && bs: 1 & \textbf{100.00} & 123.77 & 123.77 & 4.73 & 12.63  \\
         \cdashline{2-9}\noalign{\vskip 1.0ex}
         &DeepInversion & bs: 64 & 50.47 & 176.35 & 176.35 & 17.20 & 40.43  \\
         &($\downarrow2$) & bs: 1 & \textbf{100.00} & 121.94 & 121.94 & 6.30 & 16.43  \\
         \cdashline{2-9}\noalign{\vskip 1.0ex}
         & PII & bs: 21 & \textbf{100.00} & 241.54  & 241.54 & 17.53 & 38.93\\
         \cdashline{2-9}\noalign{\vskip 1.0ex}
         &\textbf{VITAL} & train-set & \underline{99.90} & \underline{58.79} & \underline{58.79} & \textbf{66.62} & \textbf{92.56}  \\
         \midrule

         % CONVNEXT
         \multirow{9}{*}{\rotatebox{90}{ConvNeXt base}}&\textit{ImageNet} &  & - & -& -& 65.66 & 89.80 \\
         \cmidrule{2-9}
         &\multirow{2}{*}{MACO} & r: 224 & 66.07  & \underline{369.64} & 62.55 & \underline{7.20} & \underline{19.77} \\
         && r: 1024 & 21.07 & 495.69 & 97.73 & 1.07 & 4.73 \\
         \cdashline{2-9}\noalign{\vskip 1.0ex}
         &\multirow{2}{*}{Fourier} & r: 224 & 60.07 & 453.91 & \underline{59.60} & 2.77 & 8.30 \\
         && r: 1024 & 14.27 & 529.33 & 77.08 & 0.60 & 2.37  \\
         \cdashline{2-9}\noalign{\vskip 1.0ex}
         &PII & bs: 16 & \textbf{100.00} & 405.50 & 92.37 & 1.97 & 6.47 \\
         \cdashline{2-9}\noalign{\vskip 1.0ex}
         &\textbf{VITAL} & &\underline{99.97} &\textbf{88.63} & \textbf{3.92} & \textbf{63.53} & \textbf{90.30}  \\
         \midrule

         % DENSENET
         \multirow{13}{*}{\rotatebox{90}{DenseNet121}}&\textit{ImageNet} &  & - & -& -& 70.64 & 93.16 \\
         \cmidrule{2-9}
         &\multirow{2}{*}{MACO} & r: 224 & 9.20 & 418.60 & 1.80 & 9.33 & 23.20 \\
         && r: 1024 & 1.60 & 475.39 & 1.98 & 1.43 & 5.03  \\
         \cdashline{2-9}\noalign{\vskip 1.0ex}
         &\multirow{2}{*}{Fourier} & r: 224 & 15.53 & 409.89 & 1.63 & 4.87 & 12.17  \\
         && r: 1024 & 1.80 & 437.18 & 1.88 & 0.90 &  3.33 \\
         \cdashline{2-9}\noalign{\vskip 1.0ex}
         &DeepInversion& bs: 64 & \textbf{100.00} & \underline{93.26} & \textbf{0.20} & 10.00 & \underline{25.47} \\
         \cdashline{2-9}\noalign{\vskip 1.0ex}
         &DeepInversion&  \multirow{2}{*}{bs: 64} &  \multirow{2}{*}{31.30} &  \multirow{2}{*}{186.16} & \multirow{2}{*}{0.83} &  \multirow{2}{*}{7.23} &  \multirow{2}{*}{20.03} \\
         & ($\downarrow$ 2) &&&&& \\
         \cdashline{2-9}\noalign{\vskip 1.0ex}
         &PII & bs: 24 & \textbf{100.00}& 377.92 & 1.23 & \underline{11.00} & 24.00  \\
         \cdashline{2-9}\noalign{\vskip 1.0ex}
         &\textbf{VITAL} & & \underline{99.93} & \textbf{79.40} & \underline{0.27} & \textbf{58.70} & \textbf{86.93 } \\
         \midrule

         % ViT-L-16
         \multirow{7}{*}{\rotatebox{90}{ViT-L-16}}&\textit{ImageNet} &  & - & -& -& 64.78  &  89.31\\
         \cmidrule{2-9}
         &MACO & r: 224 & 44.33 & 371.54 & 946.96 & 3.93 & 10.57 \\
         \cdashline{2-9}\noalign{\vskip 1.0ex}
         &Fourier & r: 224 & 25.30 & 447.56 & 990.51 & 1.67 & 5.13  \\
         \cdashline{2-9}\noalign{\vskip 1.0ex}
         &PII & bs: 2 & \textbf{100.00} & \underline{274.28} & \underline{537.32} & \underline{15.70} & \underline{32.73}   \\
         \cdashline{2-9}\noalign{\vskip 1.0ex}
         &\textbf{VITAL} & & \underline{99.80} & \textbf{128.02} & \textbf{126.29} & \textbf{68.17 }& \textbf{92.80}   \\
         \midrule

         % ViT-L-32
         \multirow{7}{*}{\rotatebox{90}{ViT-L-32}}&\textit{ImageNet} &  & - & -& -& 65.83  & 90.03 \\
         \cmidrule{2-9}
         &MACO & r: 224 & 24.87 & 280.77 & 2318.90 & 17.53 & 37.23 \\
         \cdashline{2-9}\noalign{\vskip 1.0ex}
         &Fourier & r: 224 & 17.03 & 331.86 & 1983.09 & 10.30 & 28.10  \\
         \cdashline{2-9}\noalign{\vskip 1.0ex}
         &PII & bs: 5 & \textbf{100.00}  & \underline{270.20}  & \underline{293.02} & \underline{38.47} & \underline{67.40}  \\
         \cdashline{2-9}\noalign{\vskip 1.0ex}
         &\textbf{VITAL} & & \underline{89.60} & \textbf{174.31} & \textbf{147.33} & \textbf{55.97} & \textbf{85.47}   \\

         \bottomrule
    \end{tabular}
    \caption{Comparison of methods on different architectures trained on Imagenet. We provide FID scores, CLIP Zero-shot prediction scores, and top-1 classification accuracy, indicating the \textbf{best} and \underline{second best}. In the settings, "r" indicates the resolution of the visualization, "bs" is the used batch size and indicate with ($\downarrow$ 2) the multi-resolution optimization version of DeepInversion.}
    \label{tab:supp_all}
\end{table*}

\section{Additional Results}
\label{sec:supp_experiments}

\begin{figure*}
    \centering
    \centerline{\includegraphics[width=\linewidth]
    {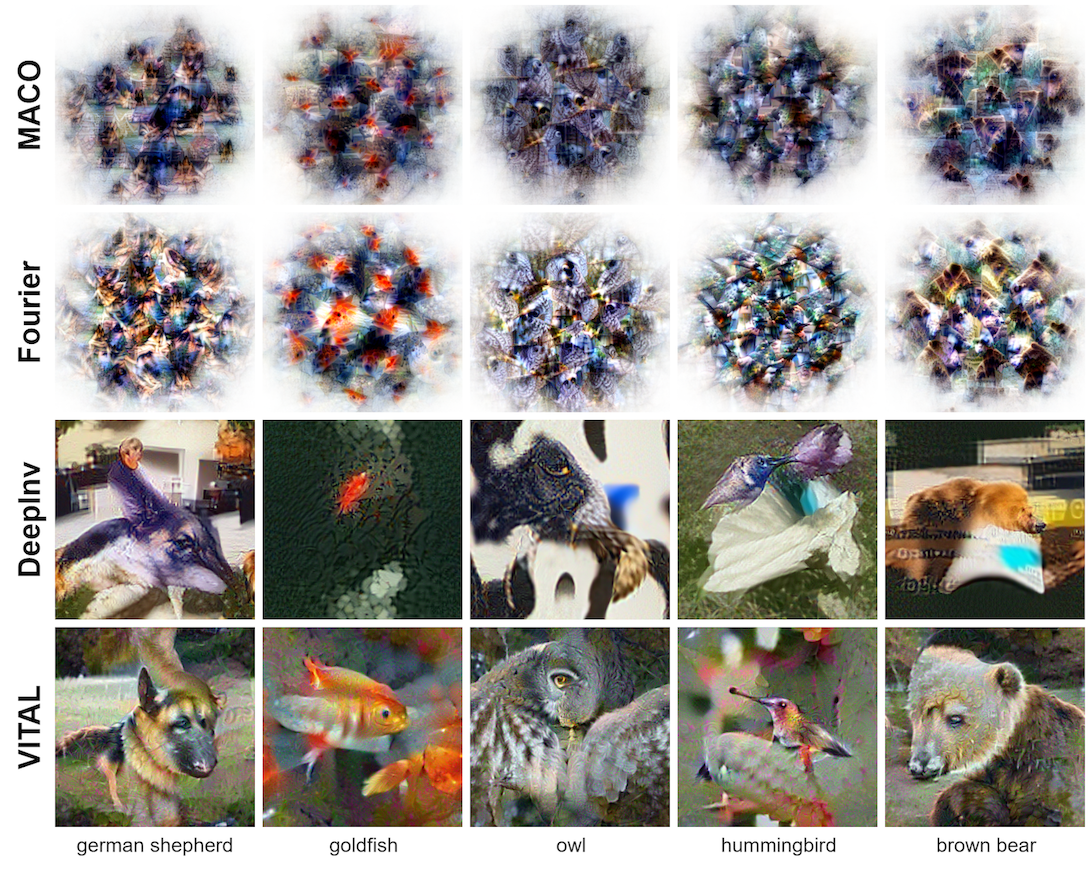}}
    \caption{\textit{Example class visualizations.} We provide more class visualizations for different classes \textbf{(columns)} of ImageNet for a trained ResNet50 model.}
    \label{fig:fig_class_neurons_1}

\end{figure*}

\begin{figure*}
    \centering
    \centerline{\includegraphics[width=\linewidth]
    {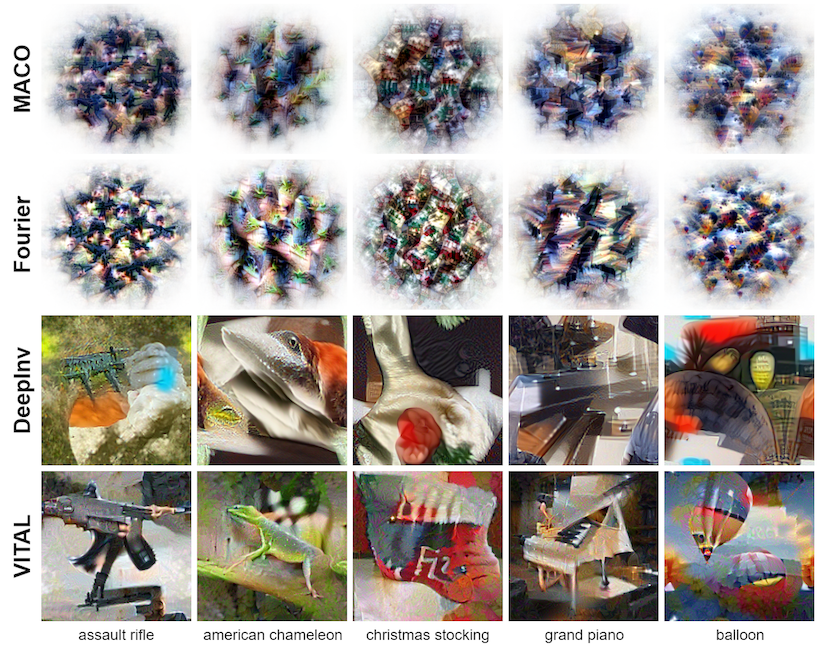}}
    \caption{\textit{Example class visualizations.} We provide more class visualizations for different classes \textbf{(columns)} of ImageNet for a trained ResNet50 model.}
    \label{fig:fig_class_neurons_2}

\end{figure*}

\begin{figure*}
    \centering
    \centerline{\includegraphics[width=\linewidth]
    {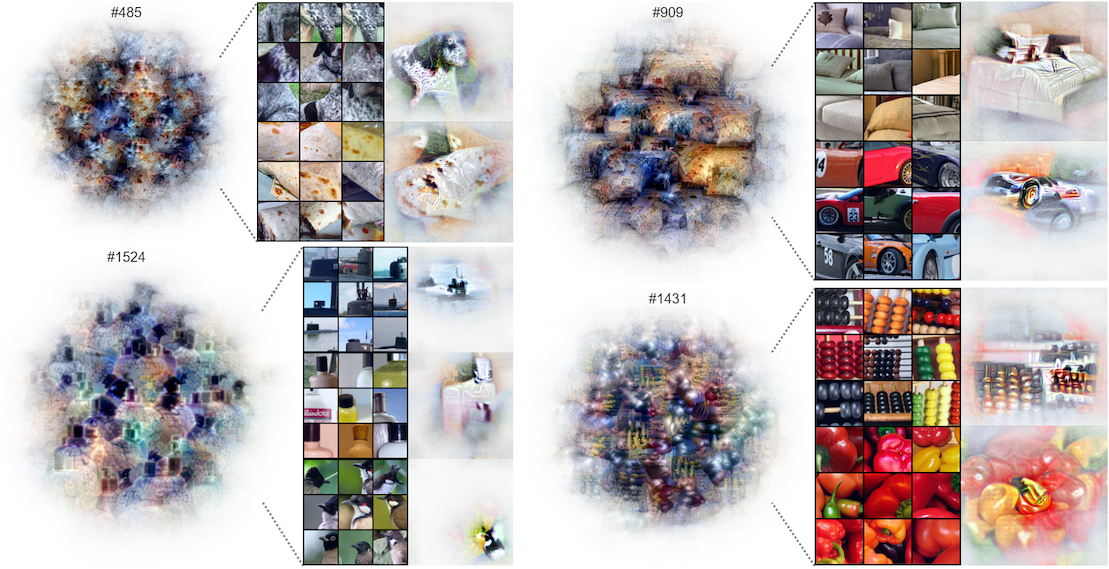}}
    \caption{\textit{Disentangling polysemanticity.} We provide four example visualizations from MACO with ResNet50 that generate visualizations that strongly activate for unrelated concepts. For each example, the first column represents the MACO visualization and the second represents the disentangled concepts from \ours. Specifically, channel \textbf{(\#485)} activates both on "burrito" and "dog body", channel \textbf{(\#909)} activates both on "mattress" and "race car", channel \textbf{(\#1524)} activates on "submarine", "lotion" and "bulbul", channel \textbf{(\#1431)} activates both on "abacus" and "bell pepper
". }
    \label{fig:fig_poly_supp}

\end{figure*}

In this section, we provide more quantitative results (\cref{tab:supp_all}, with various setups for methods Fourier \cite{olah2017feature}, MACO \cite{Fel2023}, DeepInversion \cite{yin2020dreaming}, and PII \cite{pmlr-v162-ghiasi22a}. Furthermore, we provide more qualitative results for class neurons ( \cref{fig:fig_supp_class_neurons,fig:fig_class_neurons_1,fig:fig_class_neurons_2,fig:fig_class_1_VITAL_diff,fig:fig_class_vit_l_16,fig:fig_class_vit_l_32,fig:fig_class_convnext,fig:fig_class_densenet}), intermediate neurons (\cref{fig:fig_inner_neurons_1,fig:fig_inner_neurons_2}), and results
on disentangles polysemantic neurons (\cref{fig:fig_poly_supp}). Additionally, we extend our results with the performance of intermediate neurons (see \cref{subsec:supp_performance_inner}), analysis of LRP on class neurons (see \cref{subsec:supp_analysis_lrp_class}), concept-level visualization (see \cref{subsec:supp_concept_vis}), analysis on predictions (see \cref{subsec:supp_classification}), scalability across different architectures (see \cref{subsec:supp_architectures}), and of the failure cases (see \cref{subsec:supp_failure}).

\subsection{Performance on Intermediate Neurons}
\label{subsec:supp_performance_inner}

It is essential for us to quantify the performance of intermediate neuron visualization, and AUC (Area Under the Curve) and MAD (Mean Activation Difference), as proposed in \cite{kopf2024cosy}, serve as valuable metrics for this purpose. In essence:
\begin{itemize}
    \item \textbf{AUC} measures how well a neuron’s activation distinguishes between relevant and irrelevant stimuli by computing the area under the Receiver Operating Characteristic (ROC) curve, which plots the true positive rate against the false positive rate at various threshold settings. A higher AUC indicates that a neuron more effectively captures the intended concept, demonstrating a stronger alignment between its activations and the target representation.
    \item \textbf{MAD} quantifies the difference between the mean activation of the neuron on synthetic images and the mean activation on control data points. A higher MAD suggests that the neuron responds more strongly to synthetic images compared to real ones, indicating that the synthetic stimuli successfully elicit the neuron’s preferred feature representations.
\end{itemize}
Both metrics are essential for evaluating neuron visualization: AUC assesses a neuron’s discriminative power, determining how selectively it activates for a given concept, while MAD measures how strongly a neuron responds to synthetic stimuli relative to real ones, capturing the effectiveness of the visualization method. In \cref{tab:inner_comp}, we present the average results for AUC and MAD across 90 neurons. We follow the experimental setup of \cite{kopf2024cosy}, using a control dataset composed of the top-50 real ImageNet images that most strongly activate the target neurons, while the synthetic datasets are generated with three different seeds per neuron. The results demonstrate the superiority of \ours over traditional feature visualization methods.

\subsection{Analysis of Relevance on Class Neurons}
\label{subsec:supp_analysis_lrp_class}

\begin{figure}
    \centering
    \centerline{\includegraphics[width=\linewidth]
    {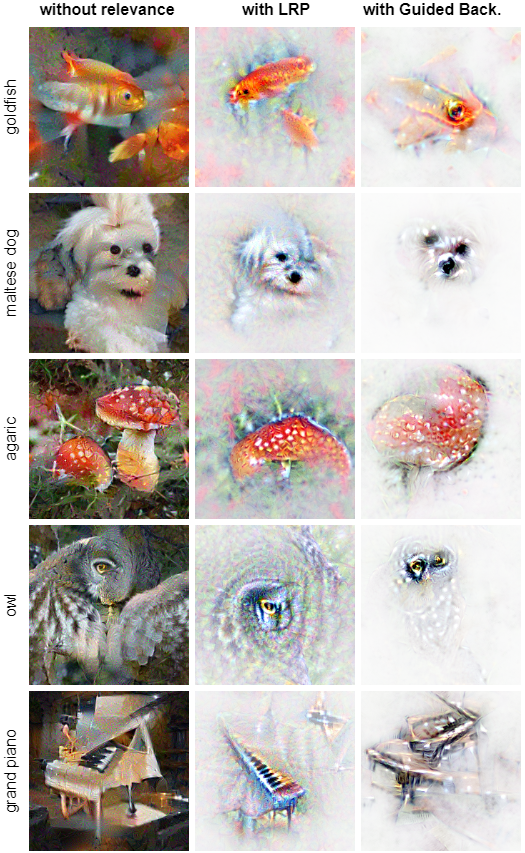}}
    \caption{\textit{Analysis on Relevance.} We performed an analysis to examine the impact of relevance information on the visualizations of class neurons on ResNet50. The findings highlight the effectiveness of LRP and Guided Backpropagation in finding the most significant regions for classification. }
    \label{fig:fig_analysis_lrp}

\end{figure}

For class neurons, we also experimented with incorporating relevance scores to factor out irrelevant activations. As for intermediate neurons, we used LRP and Guided Backpropagation to obtain the relevance maps of each building block for measuring their contribution to the final predicted class $c$ for the given $N$ images. In \cref{fig:fig_analysis_lrp}, we show a comparison between visualizing class neurons with and without relevance. As with intermediate neurons, incorporating feature relevance scores into activations and aligning their distributions would encourage background features in the FV to disappear. When comparing the visualizations of "agaric" with and without relevance, it is evident that the model focuses \textit{only} on the mushroom's cap and its spore print color for classification. For class neurons, which encapsulate an \textit{entire object}, we as humans also consider \textit{each part} of the object to understand it (e.g., stem of a mushroom). For class neurons, this hence involves a trade-off between enhancing human interpretability and maintaining \textit{faithfulness} to the model's exact reasoning mechanism, corresponding to optimization with and without relevance scores.
An interesting line of future work would be to consider LRP on self-supervised models, which usually learn more than just one distinguishing feature of an object. There, class visualization involving an attribution method would make the most sense.

\begin{table}[t]
    \centering
    \small
    \begin{tabular}{cccc}
        \toprule
         Method & Setup & AUC ($\uparrow$) & MAD ($\uparrow$) \\ 
         \midrule
         Fourier & res: 224 & 0.3073 & -0.8120\\
         MACO & res: 224 & 0.2561 & -0.9678 \\
         VITAL & & \textbf{0.5556} & \textbf{0.1587} \\
         %\midrule
         \bottomrule
    \end{tabular}
    \caption{Comparison of methods on ResNet50 trained on Imagenet for intermediate neuron visualization through AUC and MAD metrics, indication the \textbf{best}.}
    \label{tab:inner_comp}
\end{table}

\subsection{Visualization of Concepts}
\label{subsec:supp_concept_vis}
As part of Mechanistic Interpretability, people are interested in finding concept-based explanations of model behavior.
These concepts might be feature directions encoded through multiple neurons in a layer, which can be, for example, discovered by CRAFT~\cite{CRAFT}. In \ours, we obtain these directions as well as the images that highly activate the concepts through CRAFT.
To optimize for feature directions, we modify the initialization of relevances of target neurons in LRP to reflect the weights given by the feature direction. Specifically, for each image's feature map at the penultimate layer, we compute the cosine similarity with the concept direction vector. Then, we obtain the pixel location of the highest cosine similarity to assign the direction vector as the initial relevance score and apply LRP as in intermediate neuron visualization. Through this modification, \ours can give \textit{meaning} to these feature directions. We provide several examples for concept visualization in \cref{fig:fig_concepts_1,fig:fig_concepts_2,fig:fig_concepts_3,fig:fig_concepts_4}.

\begin{table}[t]
    \centering
    \footnotesize
    \begin{tabular}{cccccc}
        \toprule
         \multirow{2}{*}{Images} & \multicolumn{2}{c}{Tiger} & \multicolumn{2}{c}{Maltese dog}  \\ 
         \cmidrule{2-5}
         & \#1 score & \#2 score & \#1 score  & \#2 score \\
         \midrule
         \multirow{2}{*}{ImageNet} &  \textbf{Tiger} &  \textbf{Tiger cat} &  \textbf{Maltese dog} & \textbf{Lhasa} \\
         & 0.8625 &  0.1363 & 0.9757 & 0.0149\\
         \hdashline\noalign{\vskip 1.0ex}
         \multirow{2}{*}{MACO}  &  \textbf{Tiger} & \textbf{Apiary} & \textcolor{arrow_color}{\textbf{Silky terrier}} & \textbf{Coral reef} \\         
         & 0.5752 & 0.1616 & 0.4127 & 0.2002 \\
          \hdashline\noalign{\vskip 1.0ex}
         \multirow{2}{*}{DeepInv} & \textbf{Tiger} &  \textbf{Tiger cat} &  \textbf{Maltese dog} &  \textbf{Lhasa} \\
         & 0.9984 & 0.0013 & 0.9989 & 0.0004 \\
         \midrule
         \multirow{2}{*}{\textbf{VITAL}} & \textbf{Tiger} &  \textbf{Tiger cat} &  \textbf{Maltese dog} &  \textbf{Lhasa} \\
         & 0.8609 & 0.1382 & 0.9797 & 0.0112\\
         \bottomrule
         
    \end{tabular}
    \caption{Comparison of top-2 softmax scores of example classes across methods applied on ResNet50 trained on ImageNet. We use MACO with resolution 224 and DeepInv with batch size 64 (best setting in previous experiments) and indicate  \textcolor{arrow_color}{\textbf{misclassification}}. }
    \label{tab:accuracy}
\end{table}

\subsection{Analysis on Predictions}
\label{subsec:supp_classification}

We further investigate the predictions of images produced from different methods on ResNet50 and observed that other methods including MACO seem to produce irrelevant features that mislead the model, producing predictions unrelated to the original class. We give two examples of prediction including the second-highest class score in \cref{tab:accuracy}.

\subsection{Scalability Across Different Architectures}
\label{subsec:supp_architectures}

\begin{figure*}
    \centering
    \begin{subfigure}{0.48\linewidth}
        \centering
        \includegraphics[width=\linewidth]{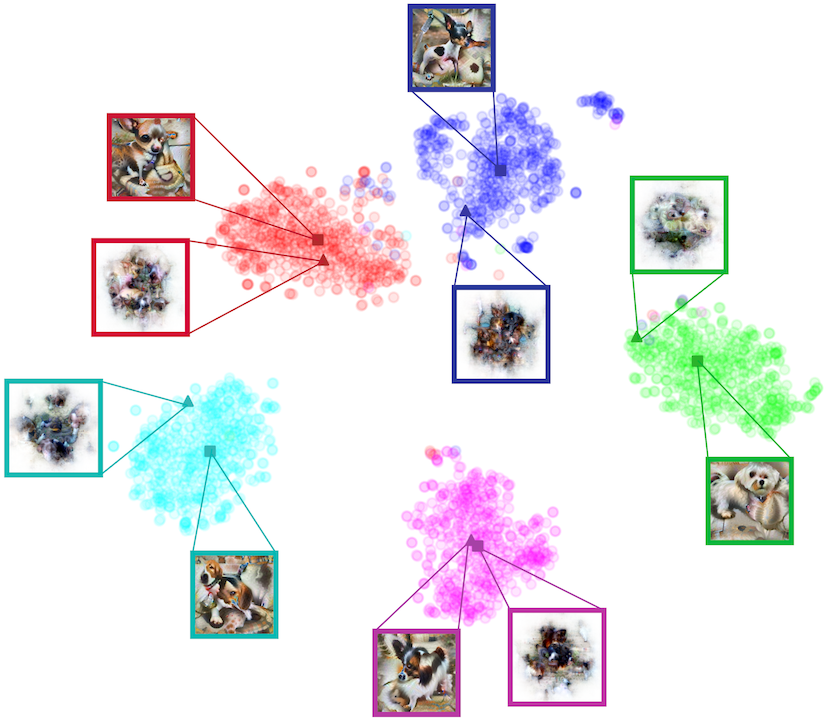}
        \caption{ConvNext-base}
    \end{subfigure}
    \hfill
    \begin{subfigure}{0.48\linewidth}
        \centering
        \includegraphics[width=\linewidth]{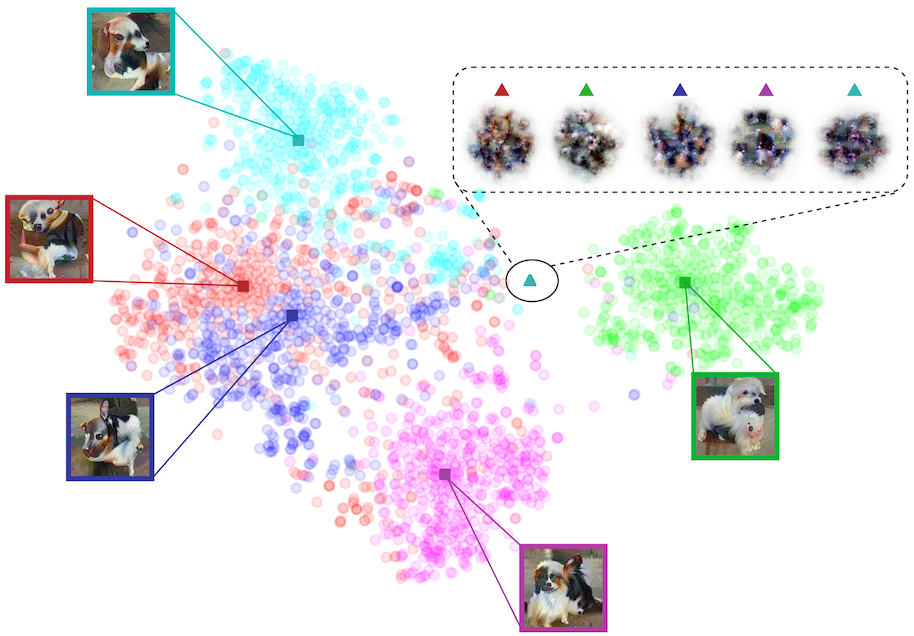}
        \caption{DenseNet121}
    \end{subfigure}
    
    \vspace{0.5cm}
    
    \begin{subfigure}{0.48\linewidth}
        \centering
        \includegraphics[width=\linewidth]{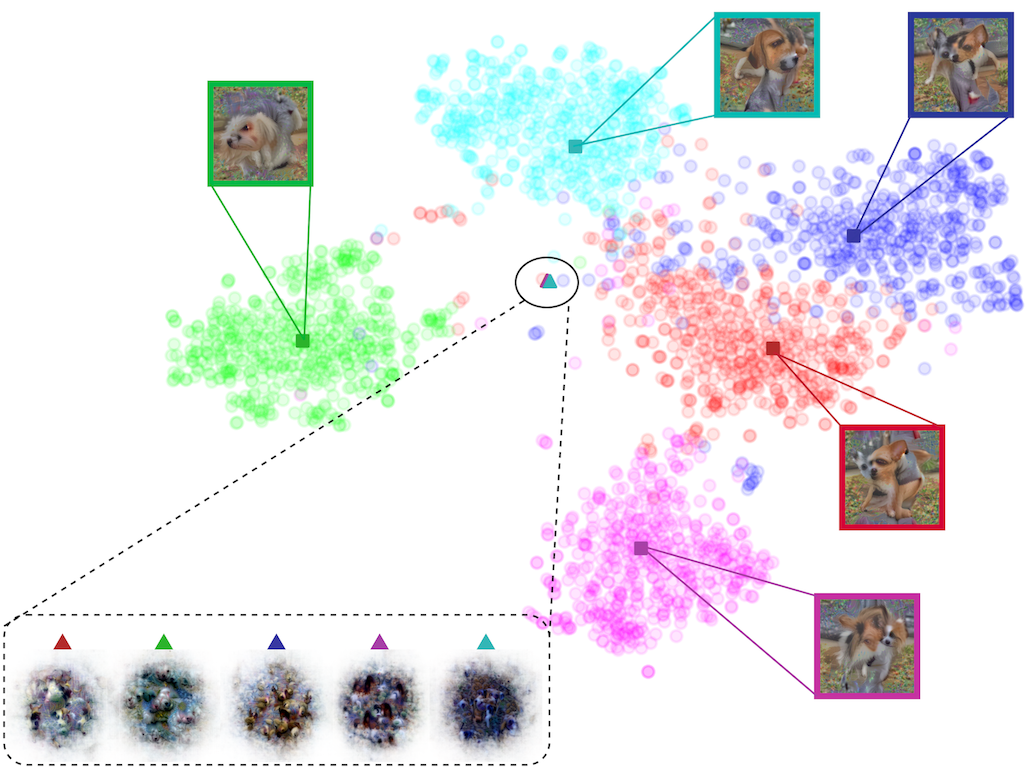}
        \caption{ViT-L-16}
    \end{subfigure}
    \hfill
    \begin{subfigure}{0.48\linewidth}
        \centering
        \includegraphics[width=\linewidth]{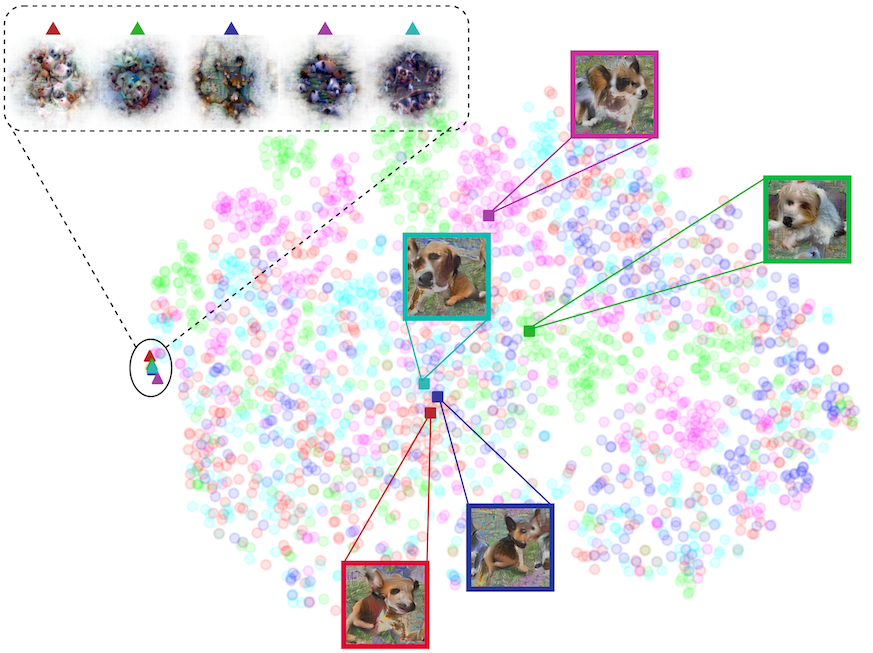}
        \caption{ViT-L-32}
    \end{subfigure}

    \caption{\textit{t-SNE projection of embedding.} We show a low-dimensional tSNE embedding of the features at the penultimate layer for five dog breeds indicated by color across different architectures. Transparent circles are original training images and FVs are indicated by symbols: \\
    $\blacksquare$: VITAL, $\blacktriangle $: MACO.}
    \label{fig:supp_tSNE}
\end{figure*}

\begin{figure*}
    \centering
    \centerline{\includegraphics[width=\linewidth]
    {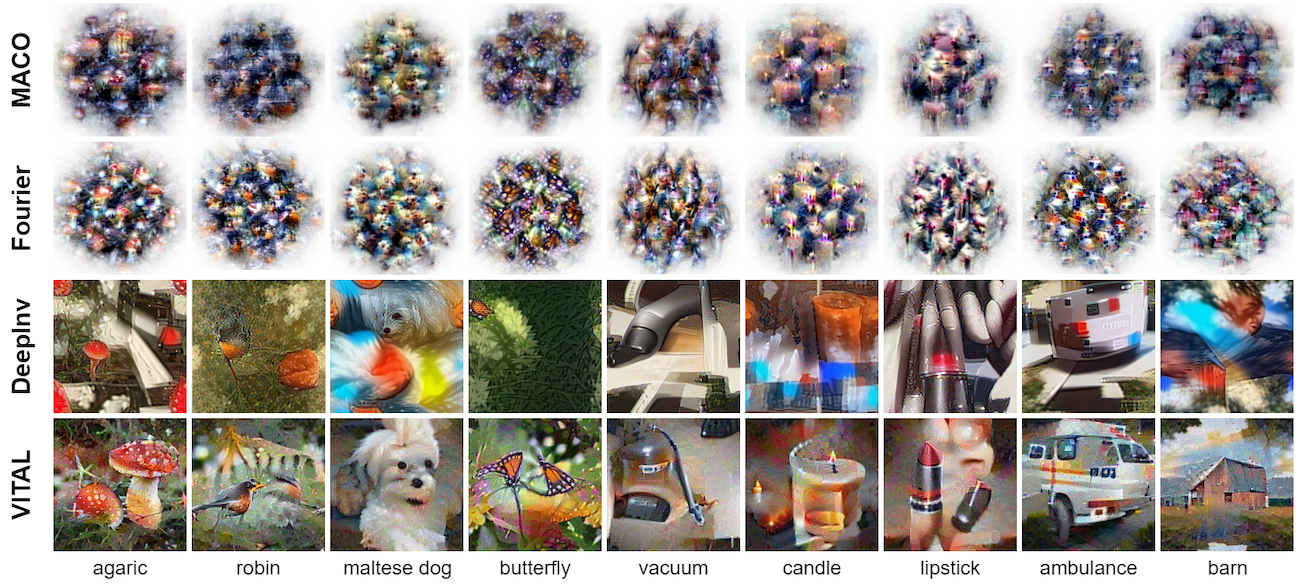}}
    \caption{\textit{Example class visualizations.} We provide class visualizations for different classes (columns) of ImageNet for a trained ResNet50 model.
    Existing work, in particular MACO and standard Fourier-based FV (top 2 rows), show highly repetitive patterns that are hard to understand.
    DeepInversion (3rd row) yields more understandable visualizations, yet suffers from artifacts that make it challenging to interpret. \ours arguably yields much more interpretable and realistic visualizations, yet, as all methods, has problems with complex spatial arrangements (see the ambulance).}
    \label{fig:fig_supp_class_neurons}

\end{figure*}

\begin{figure*}
    \centering
    \centerline{\includegraphics[width=\linewidth]
    {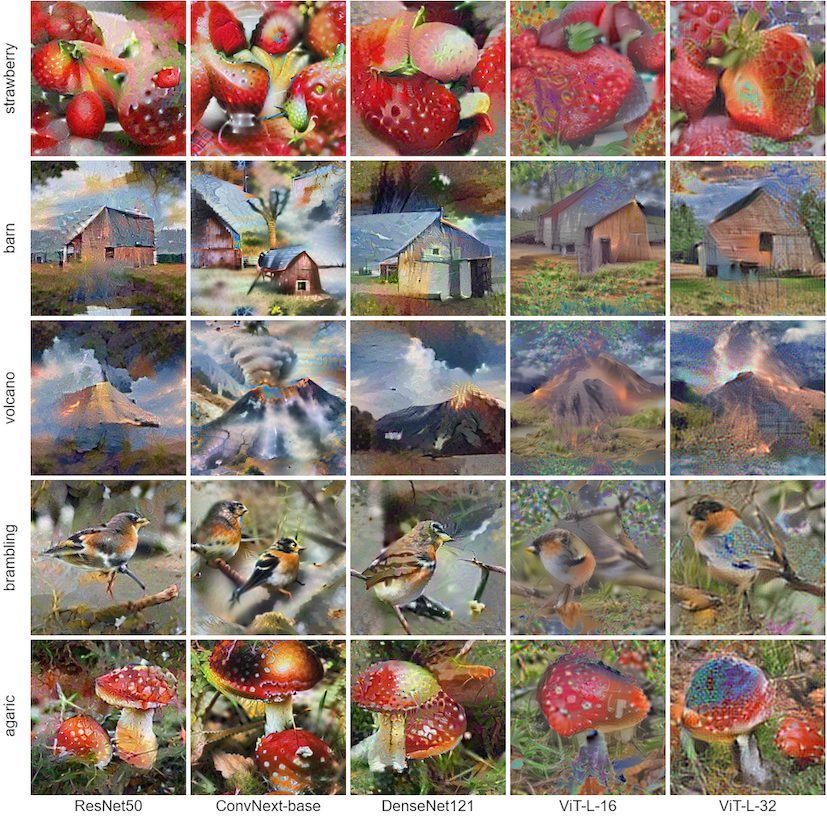}}
    \caption{\textit{Example class visualizations.} We provide more class visualizations of \ours for different classes \textbf{(rows)} of ImageNet for different models \textbf{(columns)}.}
    \label{fig:fig_class_1_VITAL_diff}

\end{figure*}

\begin{figure*}
    \centering
    \centerline{\includegraphics[width=\linewidth]
    {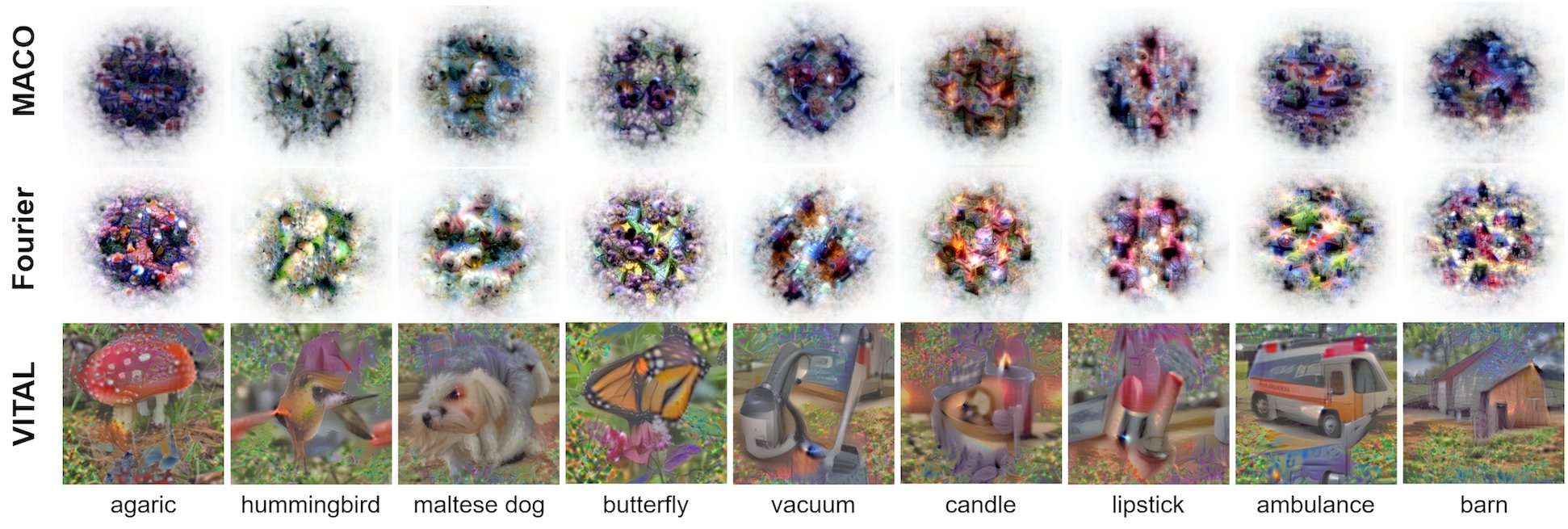}}
    \caption{\textit{Example class visualizations.} We provide more class visualizations for different classes \textbf{(columns)} of ImageNet for a trained VıT-L-16 model.}
    \label{fig:fig_class_vit_l_16}

\end{figure*}

\begin{figure*}
    \centering
    \centerline{\includegraphics[width=\linewidth]
    {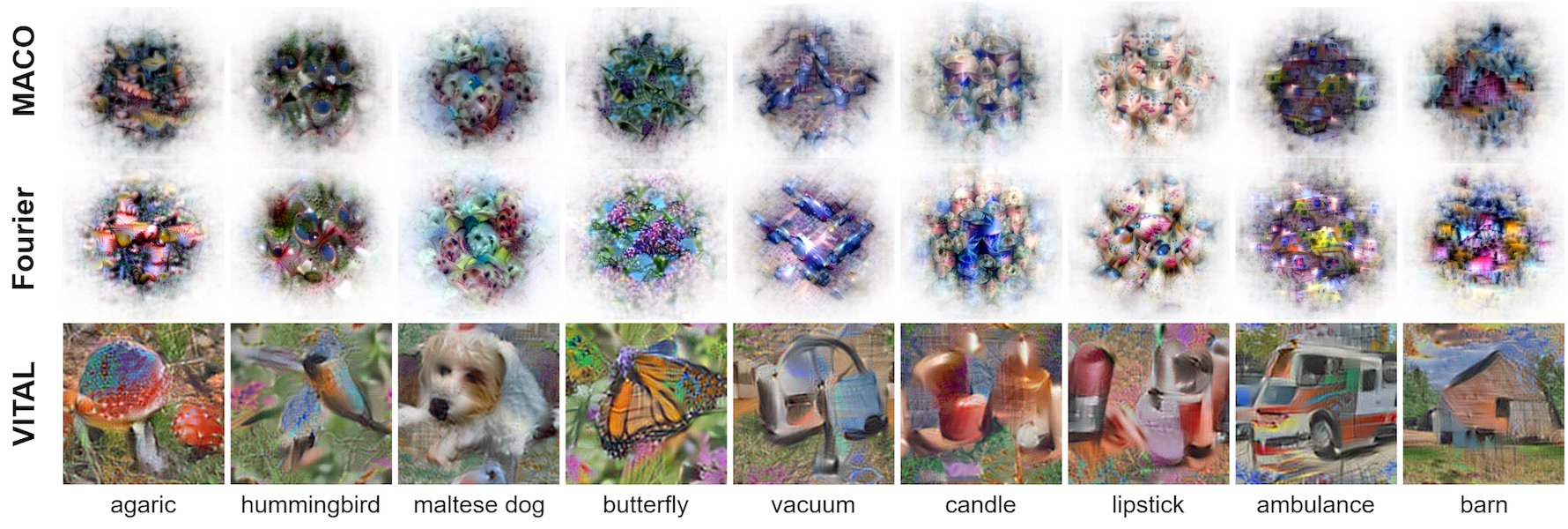}}
    \caption{\textit{Example class visualizations.} We provide more class visualizations for different classes \textbf{(columns)} of ImageNet for a trained VıT-L-32 model.}
    \label{fig:fig_class_vit_l_32}

\end{figure*}

\begin{figure*}
    \centering
    \centerline{\includegraphics[width=\linewidth]
    {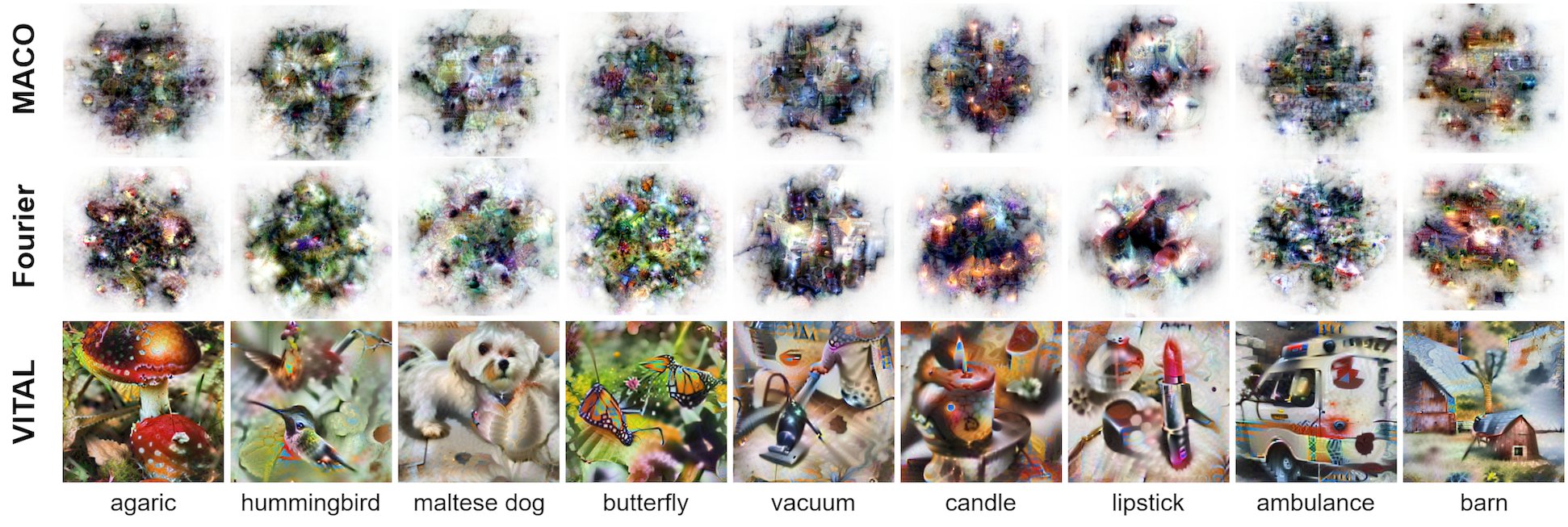}}
    \caption{\textit{Example class visualizations.} We provide more class visualizations for different classes \textbf{(columns)} of ImageNet for a trained ConvNext-base model.}
    \label{fig:fig_class_convnext}

\end{figure*}

\begin{figure*}
    \centering
    \centerline{\includegraphics[width=\linewidth]
    {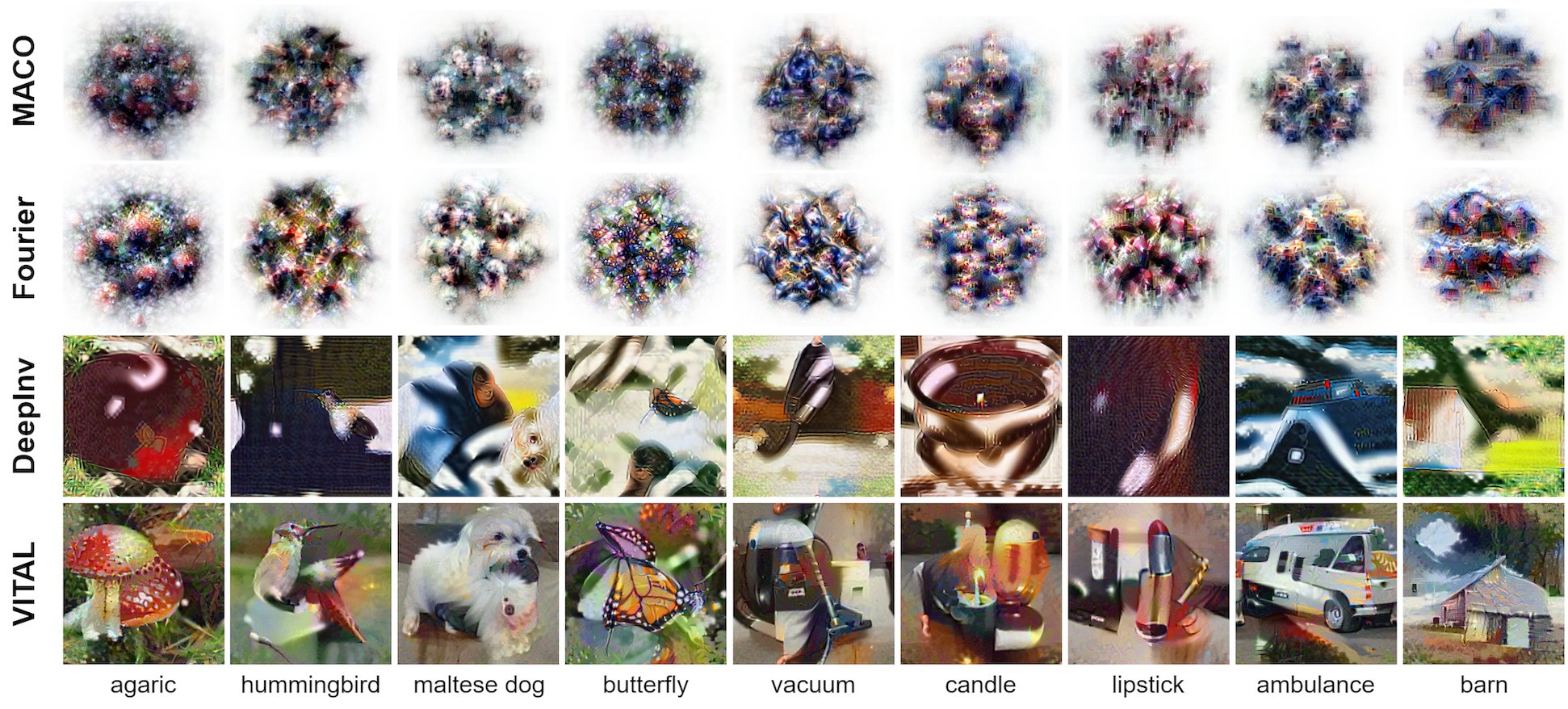}}
    \caption{\textit{Example class visualizations.} We provide more class visualizations for different classes \textbf{(columns)} of ImageNet for a trained DenseNet121 model.}
    \label{fig:fig_class_densenet}

\end{figure*}

We demonstrate the scalability of \ours by evaluating its performance across various architectures and conducting qualitative assessments (see \cref{fig:fig_supp_class_neurons,fig:fig_class_1_VITAL_diff,fig:fig_class_vit_l_16,fig:fig_class_vit_l_32,fig:fig_class_convnext,fig:fig_class_densenet}). Our results highlight the robustness of our approach, whereas other methods fail to achieve similar adaptability and consistency across different network designs. Furthermore, t-SNE projections in \cref{fig:supp_tSNE} reveal a similar trend in the embedding space. \ours is the only method that reliably position generated features at the center of their respective clusters, capturing distinct characteristics that are recognizable.

\subsection{Hardness Analysis and Failure Cases}
\label{subsec:supp_failure}

\begin{figure*}
    \centering
    \begin{subfigure}{0.48\linewidth}
        \centering
        \includegraphics[width=.9\linewidth]{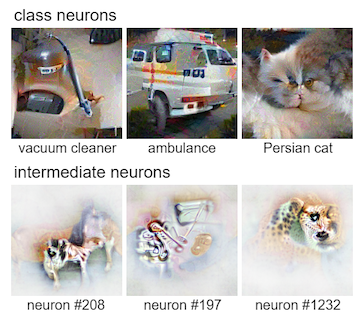}
        \caption{ResNet50}
    \end{subfigure}
    \hfill
    \begin{subfigure}{0.48\linewidth}
        \centering
        \includegraphics[width=.9\linewidth]{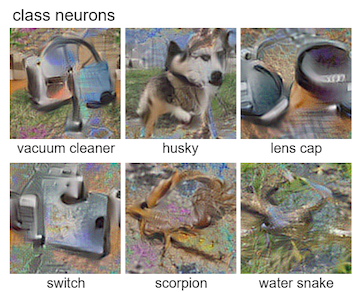}
        \caption{ViT-L-32}
    \end{subfigure}
    
    \caption{The example failure cases in visualization quality for both ResNet50 and ViT-L-32.}
    \label{fig:supp_failure}
\end{figure*}

While \ours generally produces clearer and more conceptually relevant images, there are still cases where visualization quality suffers. These negative examples highlight areas for further refinement of our framework. In \cref{fig:supp_failure} we investigate these cases, including hardness of interpretation analysis of the generated visuaizations with the help of the aferomentioned user studies. We also offer analysis based on our interpretations. We observe it is harder for people to interpret that includes particularly complex scenes, such as the \textit{vacuum cleaner} and \textit{ambulance} from ResNet50, where the concept is less distinct. Additionally, due to the distribution-matching loss, local details are lost, leading to unrealistic structure in generated images—most notably seen in the \textit{Persian cat} visualization from ResNet50 and the \textit{husky} from ViT-L-32. Holistic user studies further indicate low confidence in intermediate neuron visualizations of ResNet50 (see \cref{fig:supp_failure}), suggesting room for improvement in this area. Moreover, ViT-L-32 exhibits cases where certain classes, such as \textit{scorpion} and \textit{water snake}, are not clearly represented at all. These observations emphasize the need for a more comprehensive study of ViTs, considering the effects of individual blocks, regularization strategies, and transformation processes.

\section{Ablation Studies}
\label{subsec:supp_ablation}

\subsection{Effects of the Building Blocks on Visualization}
\label{subsec:supp_ablation_layers}

We performed an analysis of how different components of a model affect the final visualization on ResNet50. As illustrated in Fig. \ref{fig:fig_ablation_layer}, when we match the feature distribution of the coarser layers, the resulting visualization primarily captures low-level information such as colors and textures, which we refer to as the style information. As we go deeper into the network, the visualizations progressively incorporate more contextual information such as the shape or the structure of an object at the cost of increased high-frequency noise. Accordingly, we observed that we can achieve a more realistic and proper visualization result by employing all the building blocks of our model that enable us to transfer the style into the context.

\subsection{Effect of the Regularization Losses}
\label{subsec:supp_ablation_hyp}
In \cref{fig:fig_hyperparameter}, we examine the impact of the parameters, $\alpha_{\text{TV}}$ and $\alpha_{\ell_2}$, of the auxiliary regularization loss that are used to further reduce noise and small artifacts in the generated image for intermediate neuron visualization on ResNet50. It should be noted that, in \cref{fig:fig_hyperparameter}, we visualized the images without a transparency map to visualize the full extent of the impact of the regularization losses.

\subsection{Effect of the Transparency Map}
\label{subsec:supp_ablation_transparency}
Irrelevant areas of the generated image stay mostly unchanged during the optimization, essentially representing noise.
Analogous to \citet{Fel2023}, we suggest using transparency maps based on the importance of the image location during optimization to show relevant image parts only.
In brief, we accumulate the gradients of our loss across each step in the optimization.
As done in SmoothGrad \cite{SmoothGrad}, we average those gradients through the whole optimization process. 
We thus ensure the identification of the areas that have been most attended to by the network during the generation of the image. We illustrate the effect of the transparency map on ResNet50 in \cref{fig:fig_transparency}.

\begin{table}[t]
    \centering
    \footnotesize
    \begin{tabular}{l cccccc}
        \toprule
         & \multirow{2}{*}{Setup} &Acc. &  \multirow{2}{*}{FID ($\downarrow$)}& \multicolumn{2}{c}{Zero-Shot Prediction}   \\ 
         \cmidrule{5-6}%\cmidrule{6}
         & & Top1 ($\uparrow$) & & Top1 ($\uparrow$) & Top5 ($\uparrow$) \\
         \midrule
        \multirow{5}{*}{\rotatebox{90}{Vary $|\mathcal{X}_{\text{\tiny ref}}|$}} &rand5 & 99.47 & 40.30 & 57.60 & 85.87 \\
         %\hdashline\noalign{\vskip 1.0ex}
         &rand10 & 99.83 & 48.94 & 62.37 & 89.40 \\
         %\hdashline\noalign{\vskip 1.0ex}
         &rand20 & 99.87 & 54.45 & 62.90 & 90.20 \\
         &rand50 & 99.90 & 58.79 & 66.62 & 92.56 \\
         &rand100 & 99.77 & 59.28 & 65.67 & 90.50 \\
         %\midrule
         \bottomrule
    \end{tabular}
    \caption{Comparison of class-specific sampling size for the reference images on ResNet50.}
    \label{tab:xref_random}
\end{table}

\begin{figure}
    \centering
    \centerline{\includegraphics[width=\linewidth]
    {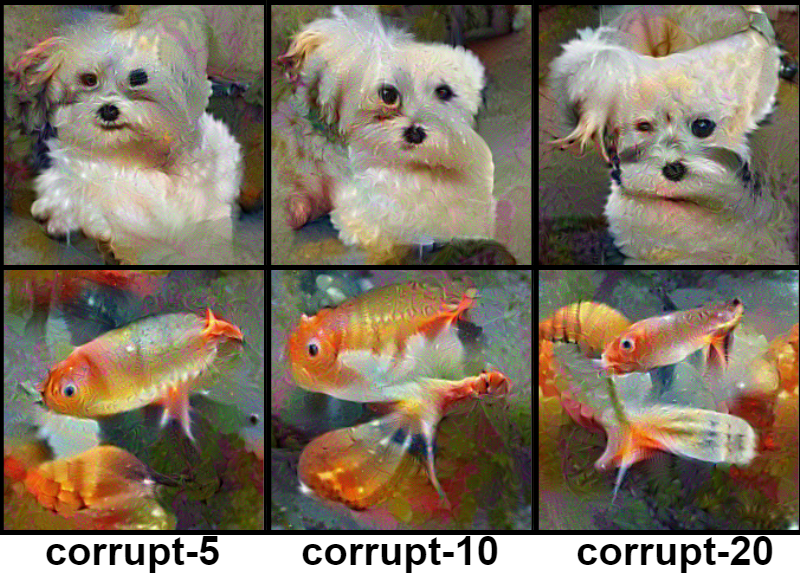}}
    \caption{The qualitative impact of corrupting the reference images on class neuron visualization.}
    \label{fig:fig_corrupt}

\end{figure}
% \begin{wraptable}{r}{0.28\textwidth}
%     \renewcommand{\tablename}{Tab.} % Local change
%     \setlength{\abovecaptionskip}{5pt}
%     \vspace{-10pt} % adjust to shift table upwards
%     \centering
%     \tiny
%     \begin{tabular}{lcccc}
%         \toprule
%         & \multicolumn{1}{c}{Acc.} & FID & \multicolumn{2}{c}{Zero-Shot Pred.} \\
%         \cmidrule(lr){4-5}
%         & Top1 $\uparrow$ & $\downarrow$ & Top1 $\uparrow$& Top5 $\uparrow$\\
%         \midrule
%         \textbf{Test Set} & 99.27 & 65.93 & 61.63 &89.30  \\
%         Train Set & 99.90 & 58.79 & 66.62 & 92.56 \\
%         Corrupt-5 & 99.97 & 72.73 & 61.03 & 88.80 \\
%         Corrupt-10 & 99.83 & 86.73 & 56.20 & 85.50 \\
%         Corrupt-20 & 99.93 & 113.30 & 47.30 & 77.40 \\
%         \midrule
%         \textbf{CIFAR-10} & - & - & 85.47 & 99.09 \\
%         VITAL &  &  & 90.00 & 100.00 \\
%         % \bottomrule
%     \end{tabular}
%     % \caption{Results}
%     \label{tab:rebuttal}
%     \vspace{-20pt} % adjust to reduce space below table
% \end{wraptable}

\begin{table}[]
    \footnotesize
    \centering
    \begin{tabular}{lcccc}
        \toprule
        & \multicolumn{1}{c}{Acc.} & FID & \multicolumn{2}{c}{Zero-Shot Prediction} \\
        \cmidrule(lr){4-5}
        & Top1 ($\uparrow$) & ($\downarrow$) & Top1 ($\uparrow$)& Top5 ($\uparrow$)\\
        \midrule
        \textbf{Test Set} & 99.27 & 65.93 & 61.63 &89.30  \\
        Train Set & 99.90 & 58.79 & 66.62 & 92.56 \\
        Corrupt-5 & 99.97 & 72.73 & 61.03 & 88.80 \\
        Corrupt-10 & 99.83 & 86.73 & 56.20 & 85.50 \\
        Corrupt-20 & 99.93 & 113.30 & 47.30 & 77.40 \\
        \midrule
        \textbf{CIFAR-10} & - & - & 85.47 & 99.09 \\
        Fourier & 5.97 & 34.87 & 11.40 & 53.80\\
        VITAL & 100.00 & 0.55 & 78.30 & 98.90\\
        
        % \bottomrule
    \end{tabular}
    \caption{The quantitative impact of corrupting the reference images, test-set analysis and CIFAR-10 analysis on class neuron visualization with ResNet50.}
    \label{tab:ref_extended}
\end{table}
\subsection{Effect of the Reference Images}
\label{subsec:supp_ablation_ref_images}

\begin{figure}
    \centering
    \centerline{\includegraphics[width=\linewidth]
    {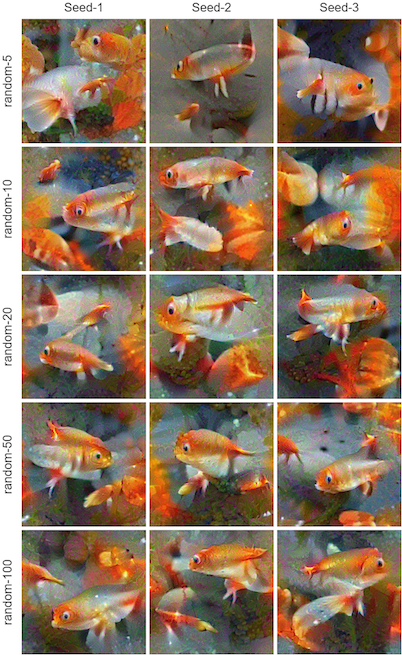}}
    \caption{The effect of class-specific reference image sampling size on class neuron visualization.}
    \label{fig:fig_Xref_random}

\end{figure}

For ResNet50, we systematically varied the Xref size for randomly selected images from a given class (see \cref{tab:xref_random} and \cref{fig:fig_Xref_random}), observing that VITAL remains robust and achieves saturation around 50 samples. This suggests that our method effectively captures the underlying feature distributions with a relatively small reference set. However, according to our preliminary analysis, when selecting random samples without considering class alignment or activation guidance, the resulting visualizations lose coherence and fail to provide meaningful insights, highlighting the importance of structured sampling in our approach.

\begin{figure*}
    \centering
    \centerline{\includegraphics[width=\linewidth]
    {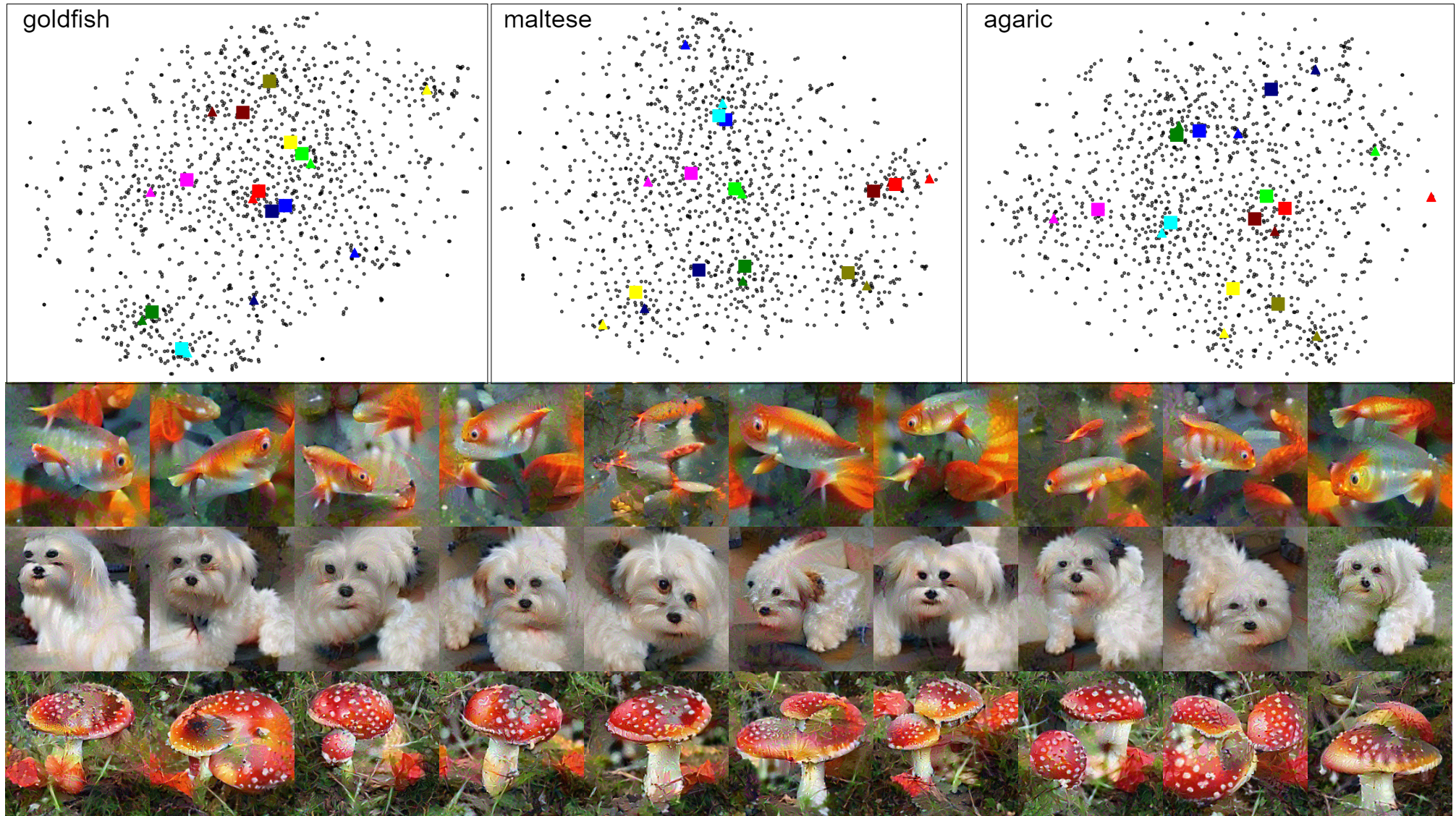}}
    \caption{VITAL visualizations generated from clustered ImageNet training samples (3 classes shown) using 50 nearest-neighbor reference images per cluster (10 clusters). $\blacksquare$: VITAL, $\blacktriangle $: cluster center.}
    \label{fig:fig_reference_images_clusters}

\end{figure*}

Additionally, we extended our embedding analysis to further validate the reference set strategy. As represented in \cref{fig:fig_reference_images_clusters}, we clustered all training samples from three ImageNet classes into subgroups and generated representative VITAL visualizations per cluster using 50 nearest neighbor images per cluster. The resulting t-SNE plots show that VITAL images cover diverse intra-class modes without collapsing to a single mode, confirming that our approach preserves both local feature fidelity and global semantic diversity within the same class. 

To verify that VITAL is not dependent on the training data and the dataset that is being used, we also consider test set examples for the reference and CIFAR-10 dataset, confirming that it can handle data beyond the original training data and the ImageNet dataset (\cref{tab:ref_extended}). Finally, we performed a corruption experiment, in which we corrupted the reference set by gradually adding 5, 10, and 20 images from outside the class (\cref{fig:fig_corrupt} and \cref{tab:ref_extended}). As expected, image quality and metrics degrade progressively with increasing contamination; however, even with partial corruption, VITAL visualizations remain considerably more stable than prior FV methods.

\subsection{Alternative Attribution Methods}
\label{subsec:supp_ablation_attribution}

\begin{figure*}
    \centering
    \centerline{\includegraphics[width=\linewidth]
    {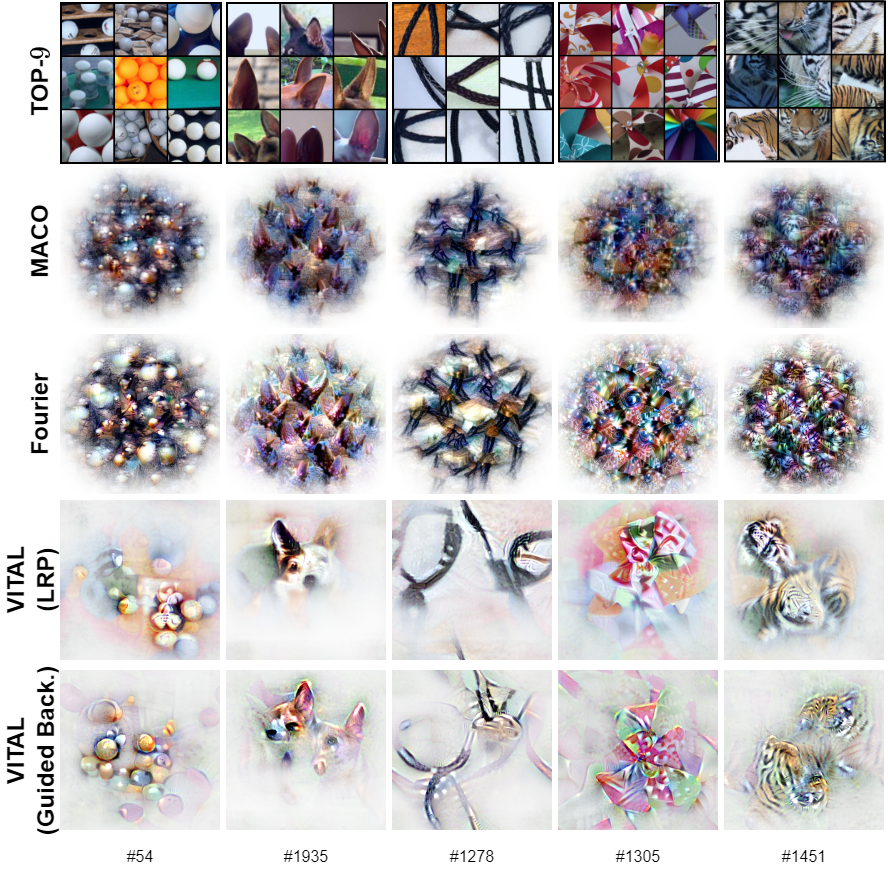}}
    \caption{\textit{Example intermediate neuron visualizations.} We provide visualizations for four randomly selected intermediate neurons \textbf{(columns)} of a trained ResNet50 model.}
    \label{fig:fig_inner_neurons_1}

\end{figure*}

\begin{figure*}
    \centering
    \centerline{\includegraphics[width=\linewidth]
    {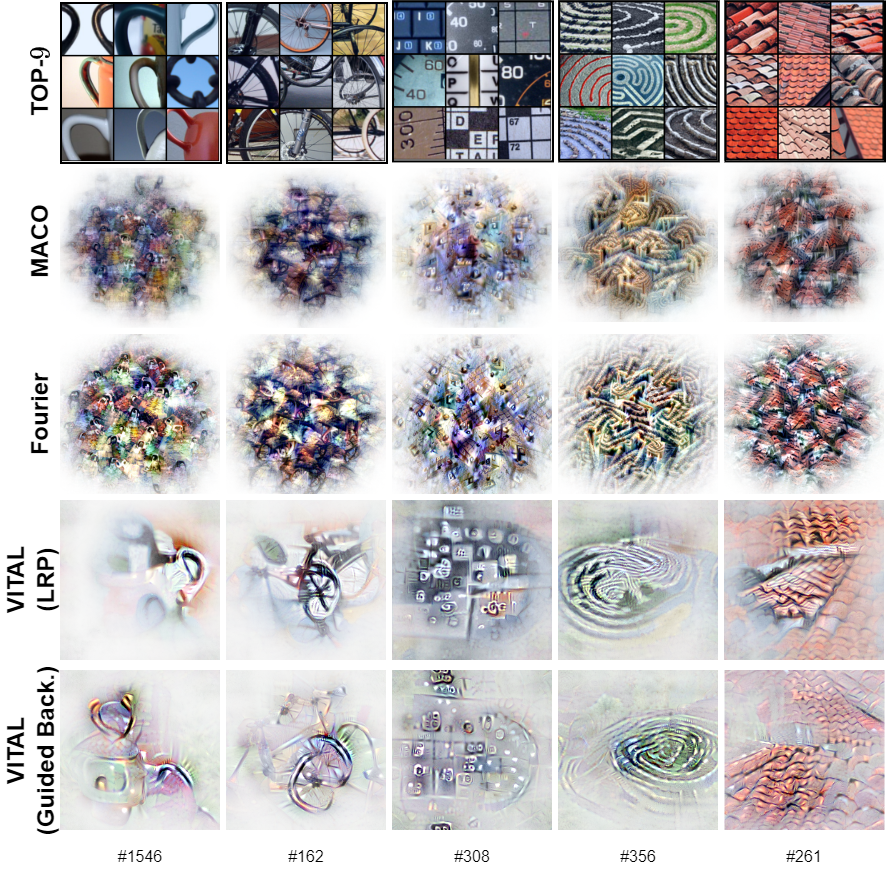}}
    \caption{\textit{Example intermediate neuron visualizations.} We provide visualizations for four randomly selected intermediate neurons \textbf{(columns)} of a trained ResNet50 model.}
    \label{fig:fig_inner_neurons_2}

\end{figure*}

\begin{figure*}
    \centering
    \centerline{\includegraphics[width=.9\linewidth]
    {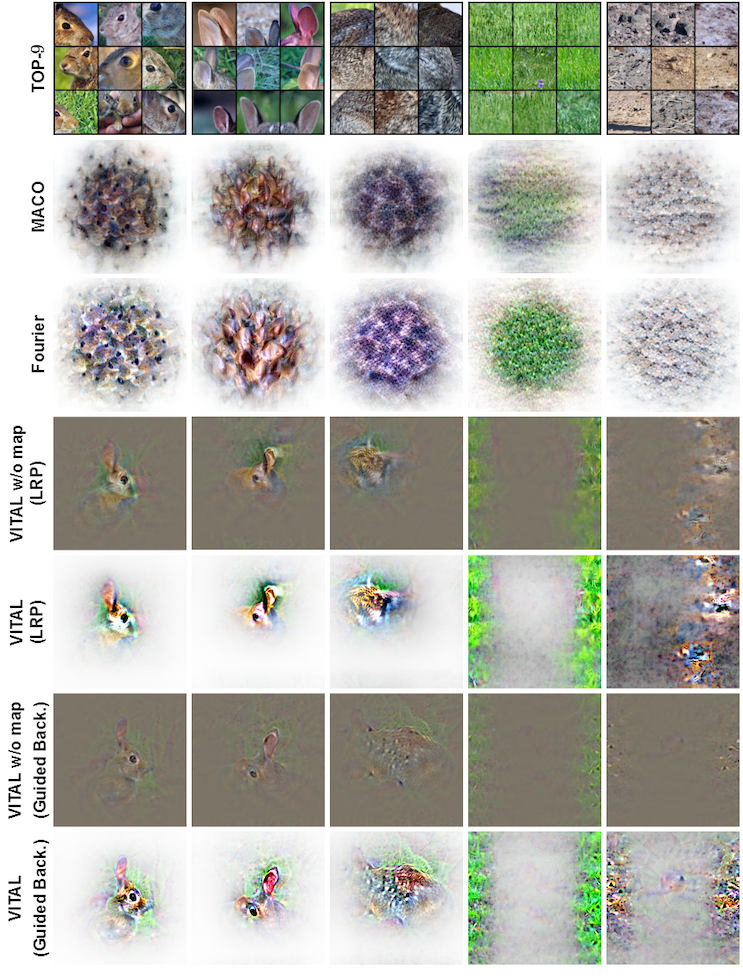}}
   \caption{\textit{Visualizing concepts.} We present example visualizations of the top five concepts identified using CRAFT for ResNet50. In this example, for the selected class \textbf{rabbit}, the top five concepts are identified as "rabbit face", "rabbit ear", "rabbit fur", "grass", and "surface".}
    \label{fig:fig_concepts_1}

\end{figure*}

\begin{figure*}
    \centering
    \centerline{\includegraphics[width=.9\linewidth]
    {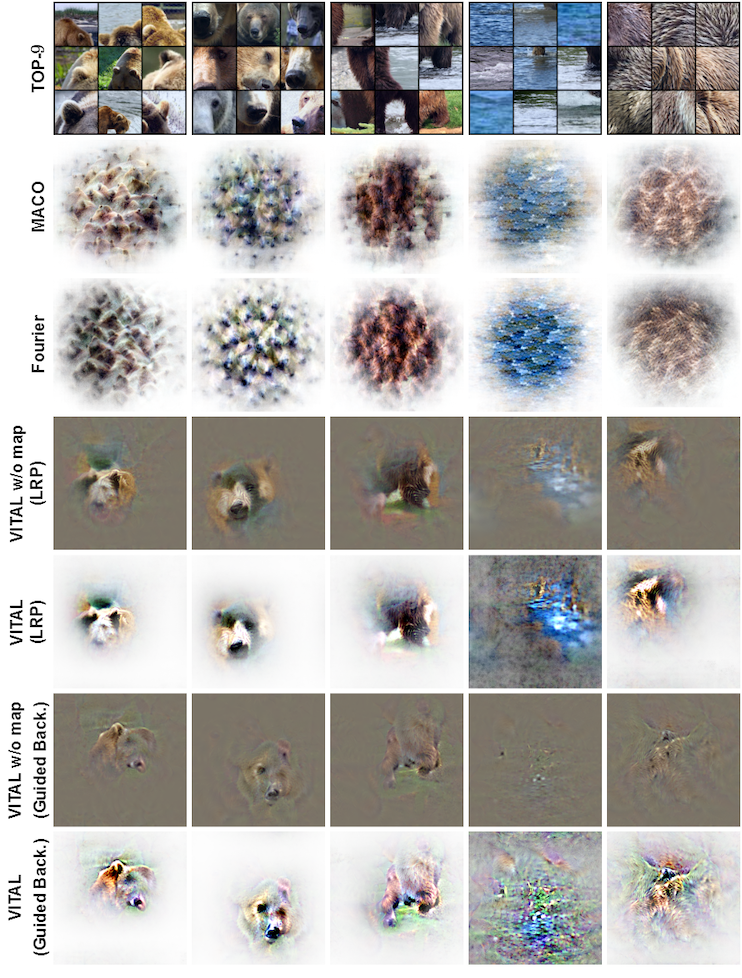}}
    \caption{\textit{Visualizing concepts.} We present example visualizations of the top five concepts identified using CRAFT for ResNet50. In this example, for the selected class \textbf{bear}, the top five concepts are identified as "bear ear", "bear face", "bear leg", "water", and "spiky fur".}
    \label{fig:fig_concepts_2}

\end{figure*}

\begin{figure*}
    \centering
    \centerline{\includegraphics[width=.9\linewidth]
    {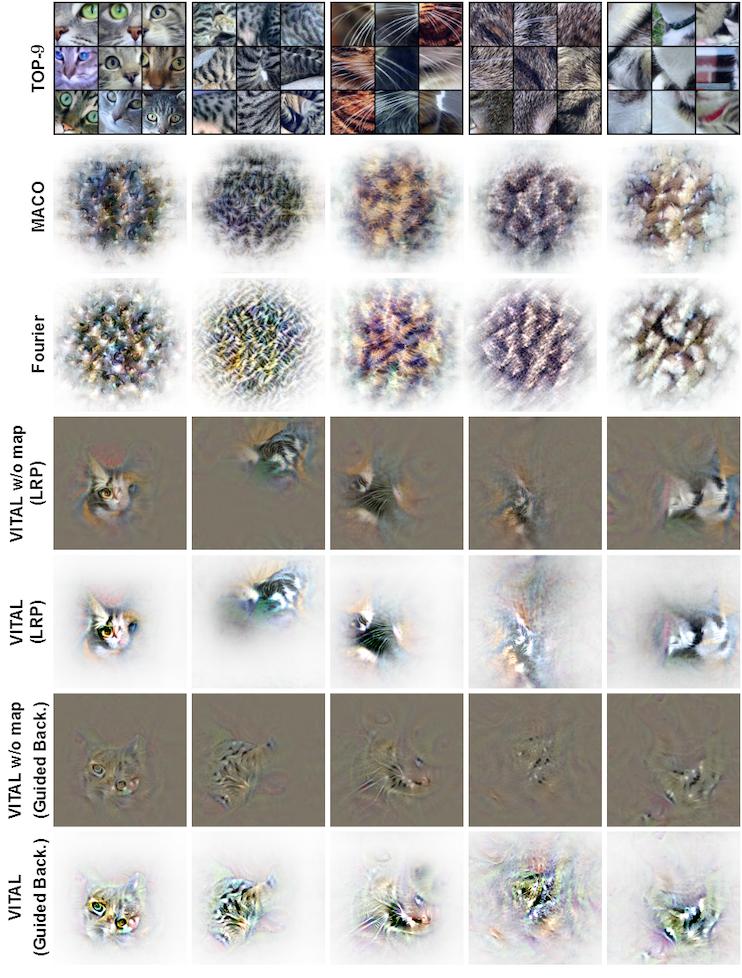}}
    \caption{\textit{Visualizing concepts.} We present example visualizations of the top five concepts identified using CRAFT for ResNet50. In this example, for the selected class \textbf{tabby cat}, the top five concepts are identified as "cat face", "fur with stripes," "cat whisker", "brown fur", and "white fur".}
    \label{fig:fig_concepts_3}

\end{figure*}

\begin{figure*}
    \centering
    \centerline{\includegraphics[width=.9\linewidth]
    {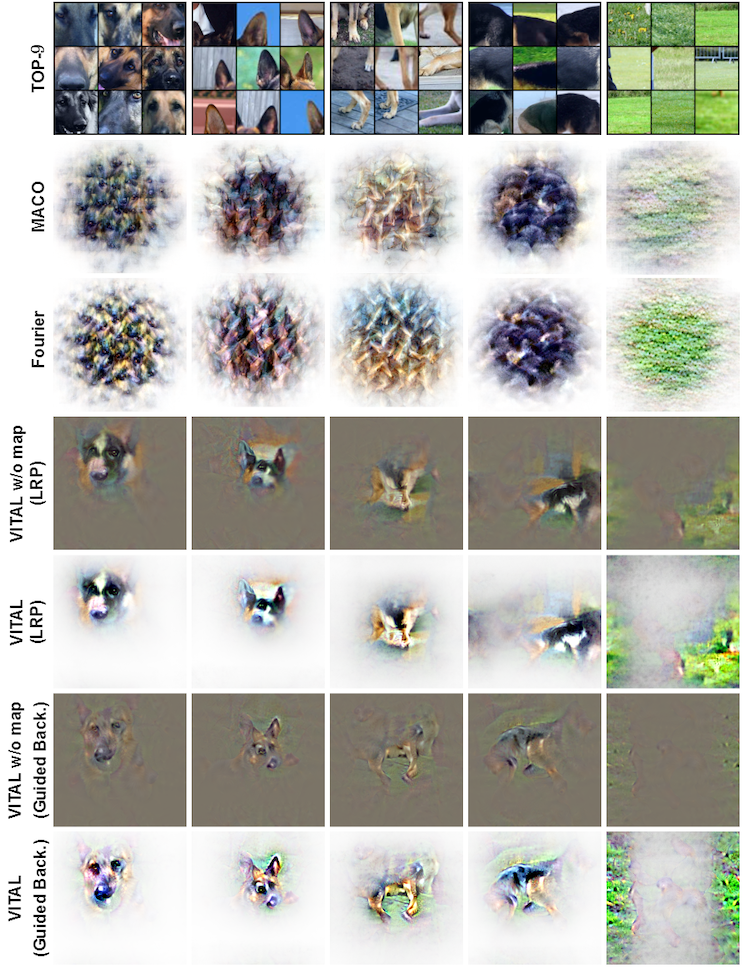}}
    \caption{\textit{Visualizing concepts.} We present example visualizations of the top five concepts identified using CRAFT for ResNet50. In this example, for the selected class \textbf{german shepherd}, the top five concepts are identified as "dog face", "dog ear", "dog leg", "dog body", and "grass".}
    \label{fig:fig_concepts_4}

\end{figure*}

We extend \ours for class neuron, intermediate neuron and concept visualizations on ResNet50 by incorporating Guided Backpropagation as an alternative to LRP, providing additional insights into feature attributions and model interpretability. We provide qualitative results to compare Guided Backpropagation with LRP for class neurons in \cref{fig:fig_analysis_lrp}, intermediate neurons in \cref{fig:fig_inner_neurons_1,fig:fig_inner_neurons_2}, and concepts in \cref{fig:fig_concepts_1,fig:fig_concepts_2,fig:fig_concepts_3,fig:fig_concepts_4}.

\begin{figure}
    \centering
    \centerline{\includegraphics[width=\linewidth]
    {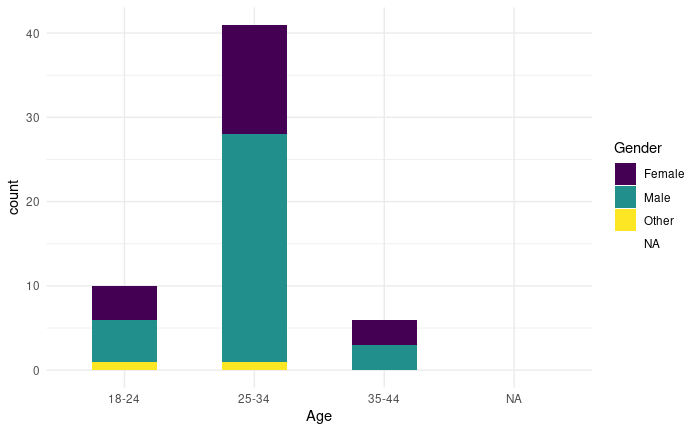}}
    \caption{\textit{Demographic Analysis.} We represent age and gender distribution of the participants in our user study.}
    \label{fig:fig_demographic}

\end{figure}

\section{Human Interpretability Study}
\label{subsec:supp_user_study}

In this section, we describe the details of our user study to quantitatively measure the performance of \ours and different FV methods on \textit{human interpretability}. 

\myparagraph{Participation.}
Participation in the study was voluntary, with 58 individuals taking part. Among these participants who disclosed their demographic information, 61.4\% identified as male, 35.1\% as female, while 3.5\% selected the option "other". Regarding age distribution within a range, 17.5\%  aged 18-24, 71.9\% aged 25-34, and 10.5 \% aged 35-44. We provide frequency distributions of the demographics with respect to age \textit{and} gender in \cref{fig:fig_demographic}. 

\begin{figure*}
    \centering
    \centerline{\includegraphics[width=.7\linewidth]
    {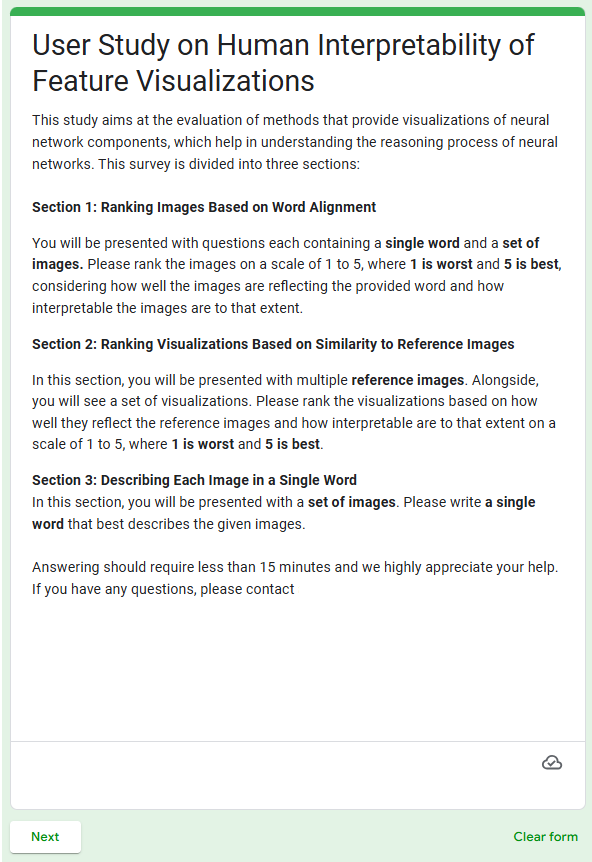}}
    \caption{\textit{Welcome page.} A screenshot of the landing page of our user study.}
    \label{fig:fig_welcome_page}

\end{figure*}

\begin{figure*}
    \centering
    \centerline{\includegraphics[width=.6\linewidth]
    {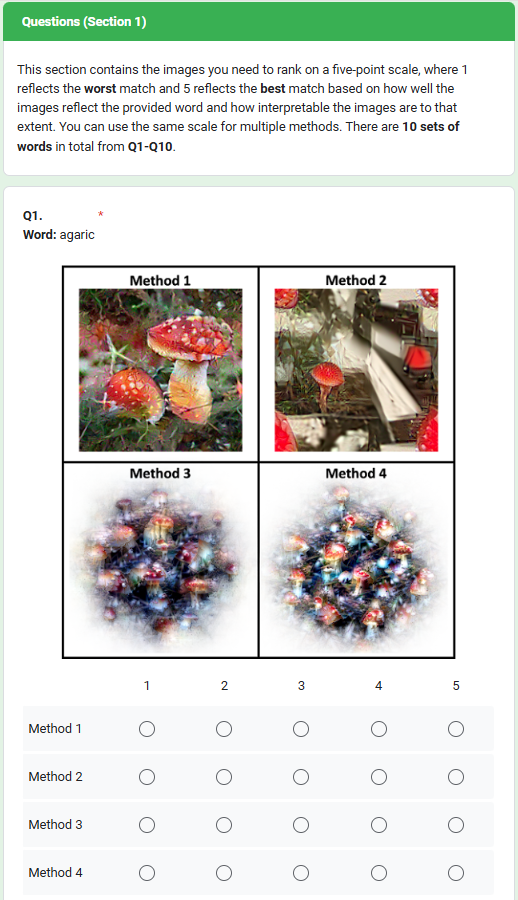}}
    \caption{\textit{Layout Section 1.} A screenshot that shows the content of section 1, including the task received with further instructions and a sample question.}
    \label{fig:fig_study_section1}

\end{figure*}

\begin{figure*}
    \centering
    \centerline{\includegraphics[width=.8\linewidth]
    {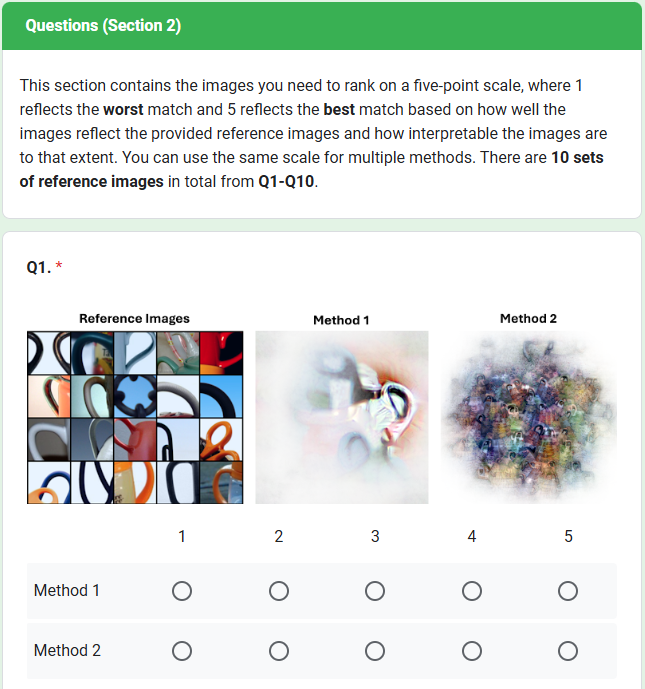}}
    \caption{\textit{Layout Section 2.} A screenshot that shows the content of section 2, including the task received with further instructions and a sample question.}
    \label{fig:fig_study_section2}

\end{figure*}

\begin{figure*}
    \centering
    \centerline{\includegraphics[width=.7\linewidth]
    {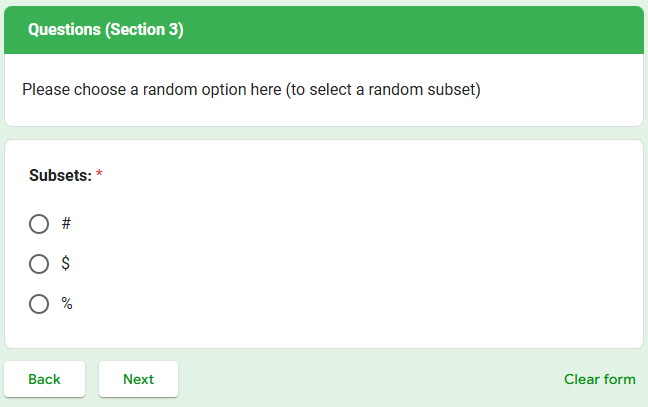}}
   \caption{\textit{Layout Section 3 subset selection.} A screenshot of the page that requires the participants to select a seed from 3 different subsets that determines the questions of section 3.}
    \label{fig:fig_study_section3_subset}

\end{figure*}

\begin{figure*}
    \centering
    \centerline{\includegraphics[width=.7\linewidth]
    {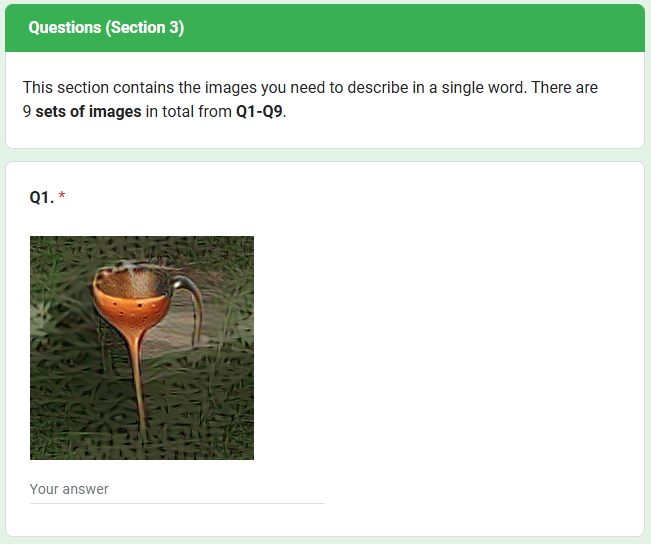}}
    \caption{\textit{Layout Section 3.} A screenshot that shows the content of section 3, including the task received with further instructions and a sample question.}
    \label{fig:fig_study_section3}

\end{figure*}

\begin{figure*}
    \centering
    \centerline{\includegraphics[width=.7\linewidth]
    {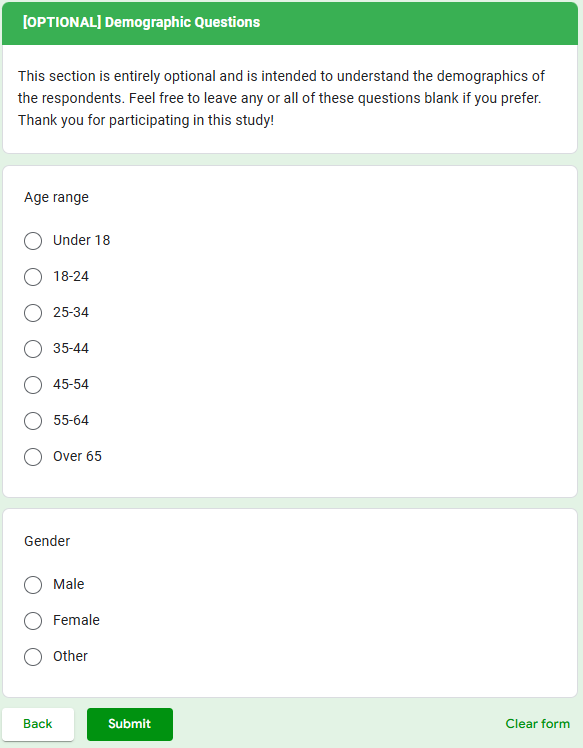}}
    \caption{\textit{Layout demographic questions.} A screenshot that shows the (optional) questions on age and gender.}
    \label{fig:fig_study_demographic}

\end{figure*}

\myparagraph{Study layout.} The study design is described in the main paper, we will here describe the layout of the three parts of the study. We conducted the user study in Google Forms. Participants were initially redirected to a welcome page, where the study's general purpose and procedures were clearly explained (see \cref{fig:fig_welcome_page}). Subsequently, they were presented with the first section of our user study, where given a single word, they evaluated how well a FV reflects the provided word. This section contains 10 sets of words and FVs in total numbered Q1-Q10 with a simple scoring system from 1 (worst) to 5 (best) to rank the visualizations (see \cref{fig:fig_study_section1}). In the second section of our user study, users evaluated how well the FVs for an inner neuron reflect the provided reference images (highly activating on the target neuron). This section contains 10 sets of reference images in total numbered Q1-Q10 with a simple scoring system from 1 (worst) to 5 (best) to rank the visualizations (see \cref{fig:fig_study_section2}). For section 3, participants were first asked to select one of three subsets (see \cref{fig:fig_study_section3_subset}), with each subset consisting of 9 questions from Q1-Q9 that required them to describe a given generated image with a word or a short description (see \cref{fig:fig_study_section3}).
To ensure comparability of methods, in each question corresponding to a given target word, each subset had one specific methods' visualization for that question. For example, in Q1, the target class was Espresso, and subset 1 had a FV of \ours, subset 2 a FV of DeepInversion, and subset 3 a FV of MACO.
Finally, participants were presented with an optional section on demographic analysis (see \cref{fig:fig_study_demographic}) before submitting the user study.

\begin{figure*}
    \centering
    \centerline{\includegraphics[width=.8\linewidth]
    {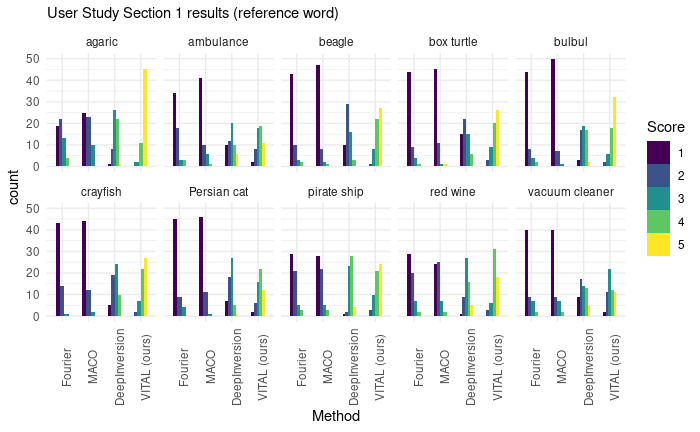}}
    \caption{The statistics on the scores for the different methods obtained for the first part of our user study, separated by class.}
    \label{fig:fig_study1_perclass}

\end{figure*}

\begin{figure*}
    \centering
    \centerline{\includegraphics[width=.8\linewidth]
    {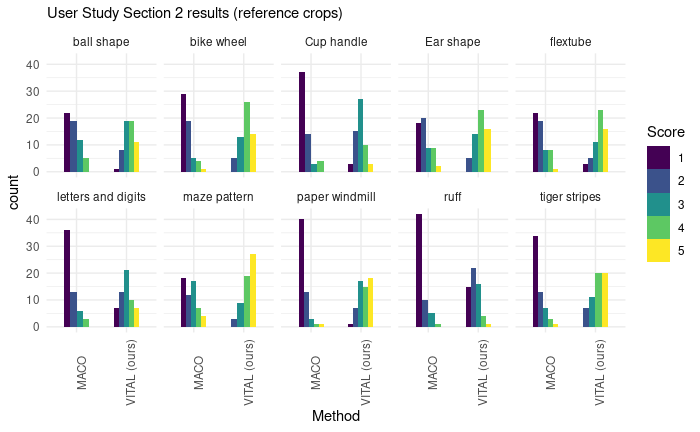}}
    \caption{The statistics on the scores for the different methods obtained for the second part of our user study, separated by concept.}
    \label{fig:fig_study2_perconcept}

\end{figure*}

\begin{figure*}
    \centering
    \centerline{\includegraphics[width=.8\linewidth]
    {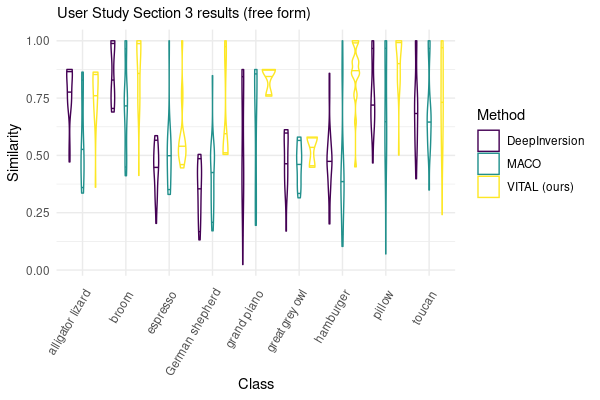}}
    \caption{The violin plots with median, 5\% and 95\% quantiles of the achieved similarity for the last part of our user study, separated by class.}
    \label{fig:fig_study3_perclass}

\end{figure*}

\myparagraph{Analysis and Results.}
In \cref{fig:fig_study1_perclass,fig:fig_study2_perconcept,fig:fig_study3_perclass}, we provide a fine-grained analysis of each question across all sections of our user study complementing the results in the main paper.
In particular, we provide class- or concept-specific score distribution for each method.
We observe that, as before, our method performs favorably across all three tasks compared to other methods for each question. Furthermore, we see that \ours yields consistently good results, showing better results in almost all cases across all study sections. There are specific classes, such as specific animals, or "hamburger" and "grand piano", where our method yields much more interpretable visualizations.

\begin{figure*}
    \centering
    \begin{subfigure}{\linewidth}
        \centering
        \includegraphics[width=.6\linewidth]{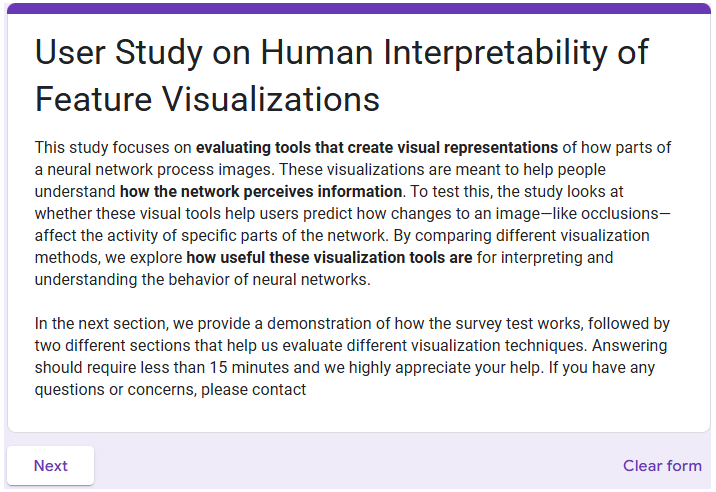}
        \caption{Welcome Page}
    \end{subfigure}
    \vspace{0.5cm}
    \begin{subfigure}{0.48\linewidth}
        \centering
        \includegraphics[width=\linewidth]{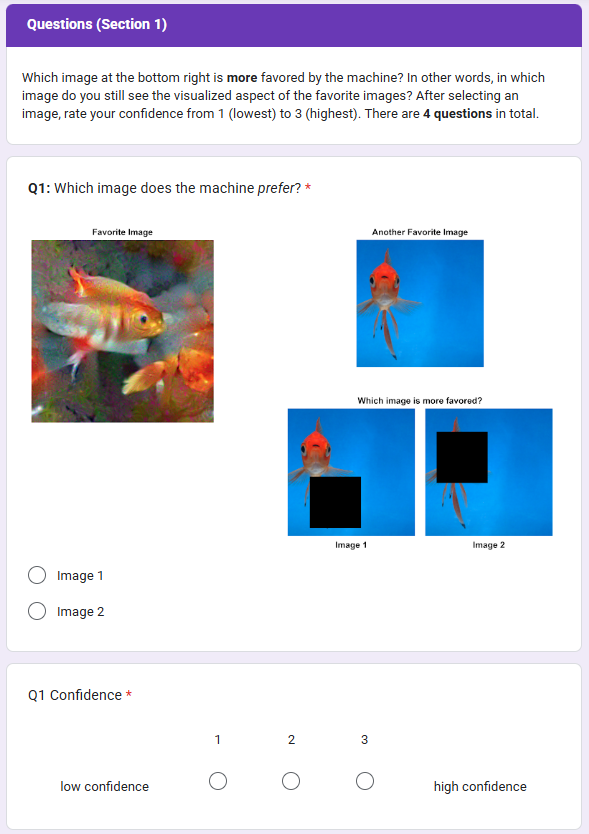}
        \caption{Section-1}
    \end{subfigure}
    \hfill
    \begin{subfigure}{0.48\linewidth}
        \centering
        \includegraphics[width=\linewidth]{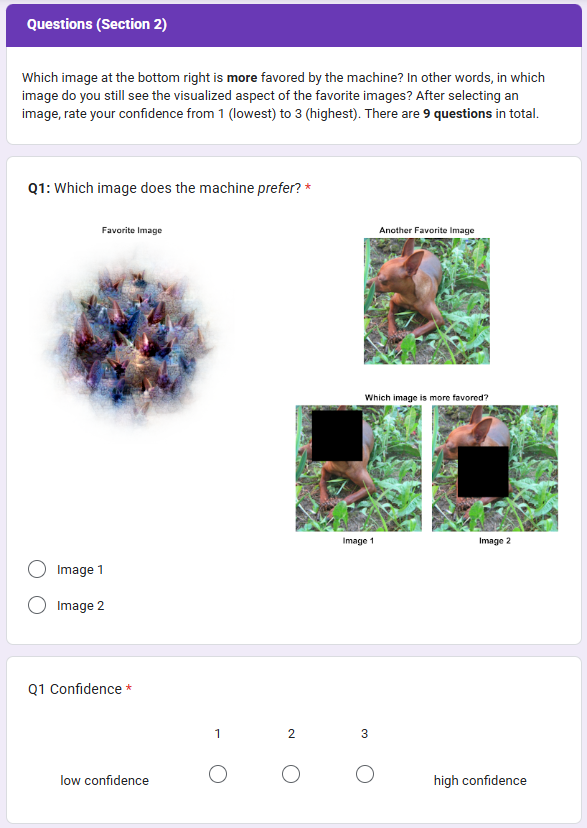}
        \caption{Section-2}
    \end{subfigure}

    \caption{Layout of the validated user study from \cite{Fel2023, zimmermann2021causal}, including the welcome page and example questions from section-1 and section-2.}
    \label{fig:supp_holistic_study}
\end{figure*}

\begin{figure*}
    \centering
    \begin{subfigure}{.98\linewidth}
        \centering
        \includegraphics[width=.55\linewidth]{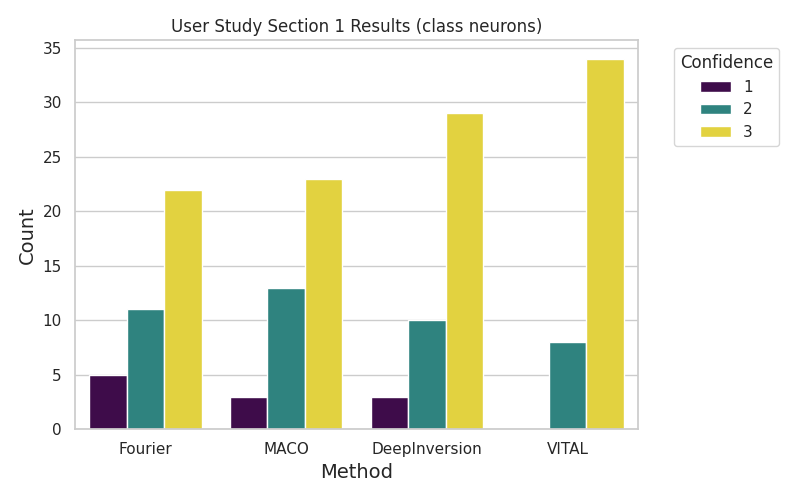}
        \caption{Section-1 Summary}
    \end{subfigure}
    \vspace{0.5cm}
    \begin{subfigure}{.98\linewidth}
        \centering
        \includegraphics[width=.65\linewidth]{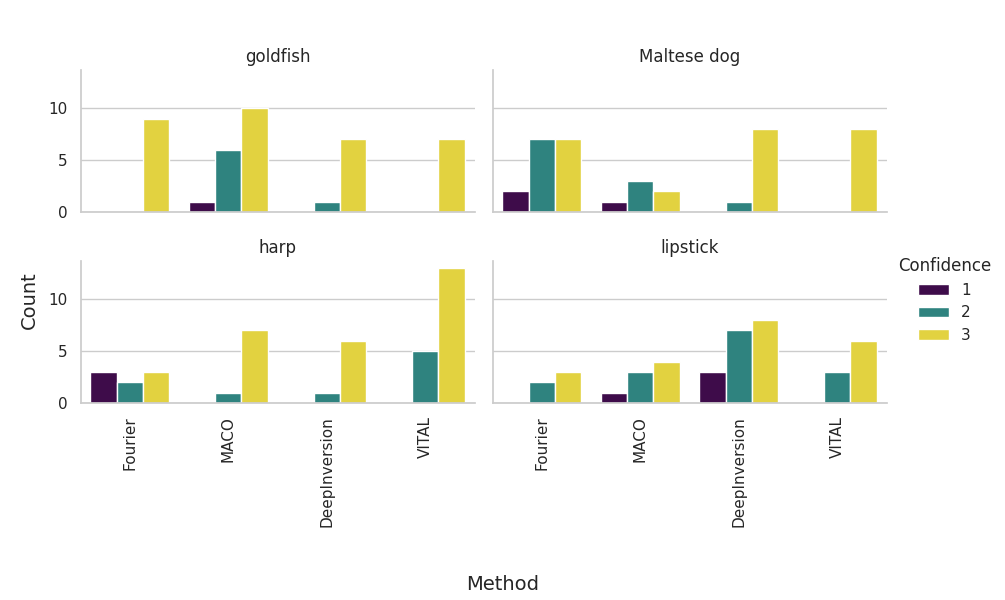}
        \caption{Section-1 Class-Specific}
    \end{subfigure}
    \vspace{0.5cm}
    \begin{subfigure}{.98\linewidth}
        \centering
        \includegraphics[width=.55\linewidth]{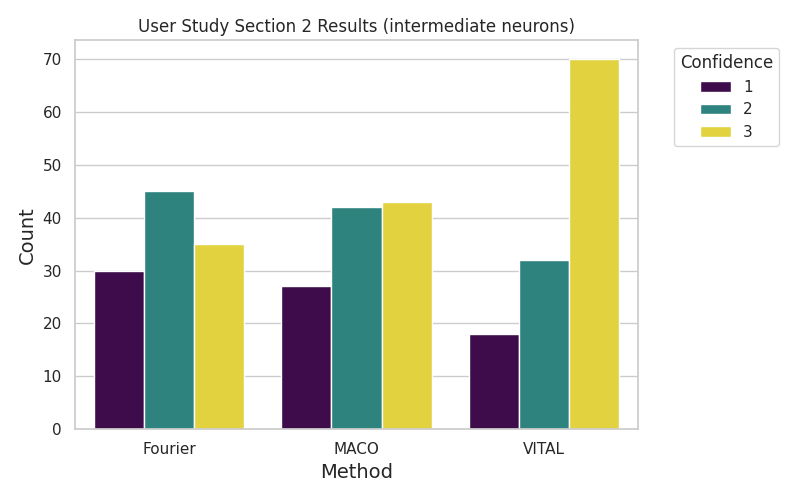}
        \caption{Section-2 Summary}
    \end{subfigure}

    \caption{The statistics on the scores for the different methods obtained for the holistic user study \cite{zimmermann2021causal, Fel2023}.}
    \label{fig:supp_holistic_study_results}
\end{figure*}

\myparagraph{Holistic User Evaluation}
As a complement to our proposed user study, we conducted a validated user evaluation following the protocol from \cite{zimmermann2021causal, Fel2023}. In this study, we recruited $N=42$ participants and replicated the setup using four randomly selected class neurons and nine randomly selected inner neurons from ResNet50. The user study is composed of two sections, where in section 1, we incorporated four different class neurons with 4 subset of questions, and in section 2, we incorporated 9 different intermediate neurons with 3 subset of questions. We have included a demonstration section to enhance clarity of the study. We provide layout snapshots from the user study in \cref{fig:supp_holistic_study}. We measured correctness based on participant confidence (maximum score of 3). For class neurons, VITAL achieved 100\% correctness (2.81), outperforming MACO at 92.86\% (2.51), Fourier at 90.48\% (2.45), and DeepInv at 100\% (2.62). Similarly, for intermediate neurons, VITAL demonstrated superior performance with 95.24\% correctness (2.43), compared to MACO at 88.89\% (2.14) and Fourier at 87.30\% (2.05). We provide fine grained analysis in \cref{fig:supp_holistic_study_results} for both section-1 and section-2. These results confirm that VITAL outperforms existing feature visualization (FV) methods in supporting interpretability.

\clearpage


\end{document}